\documentclass[]{bytedance_seed}

\usepackage[toc,page,header]{appendix}

\usepackage{xcolor}
\usepackage{minitoc}
\usepackage{tcolorbox}  
\usepackage{fontawesome5}
\usepackage{pifont}
\usepackage{booktabs}      
\usepackage{array}         
\usepackage{colortbl}      
\usepackage{graphicx}      
\usepackage{caption}       
\usepackage{amsmath}       
\usepackage{xspace}     
\usepackage{booktabs}   
\usepackage{enumitem}
\usepackage{makecell} 
\usepackage{amsmath, amssymb} 

\definecolor{boxcolor}{HTML}{d92523} 
\definecolor{bulbcolor}{HTML}{e3b87f} 

\newcommand{\benchmark}{\textit{Encyclo-K}\xspace}

\title{\benchmark: Evaluating LLMs with Dynamically Composed Knowledge Statements}

\author[1,2,5*]{Yiming Liang}
\author[5,7*]{Yizhi Li}
\author[3\dagger]{Yantao Du}
\author[3,5]{Ge Zhang}
\author[5]{Jiayi Zhou}
\author[3]{Yuchen Wu}
\author[3]{Yinzhu Piao}
\author[3]{Denghui Cao}
\author[3]{Tong Sun}
\author[3]{Ziniu Li}
\author[6]{Li Du}
\author[6]{Bo Lei}
\author[4,5]{Jiaheng Liu}
\author[7]{Chenghua Lin}
\author[4]{Zhaoxiang Zhang}
\author[3\dagger]{Wenhao Huang}
\author[1,2\dagger]{Jiajun Zhang}

\affiliation[1]{School of Artificial Intelligence, University of Chinese Academy of Sciences}
\affiliation[2]{Institute of Automation, Chinese Academy of Sciences}
\affiliation[3]{Bytedance Seed China}
\affiliation[4]{Nanjing University}
\affiliation[5]{M-A-P}
\affiliation[6]{BAAI}
\affiliation[7]{The University of Manchester}

\contribution[*]{Equal contribution}
\contribution[\dagger]{Corresponding authors}

\abstract{
Robust benchmarks are essential for accurately reflecting the generalization capabilities of large language models (LLMs).
Existing benchmarks that curate questions at the question level suffer from three limitations: vulnerability to data contamination, restriction to single-concept assessment, and reliance on costly domain expert annotation.
We propose \benchmark, a statement-based benchmark that extracts standalone knowledge statements from authoritative textbooks and dynamically composes them into evaluation questions through random sampling at test time.
This design directly addresses all three limitations: the combinatorial space resists memorization while maintaining stable model rankings across question sets; each question aggregates 8--10 statements for comprehensive knowledge assessment; and annotators only verify formatting compliance without requiring domain expertise.
Experiments on over 50 LLMs demonstrate that \benchmark poses substantial challenges—even OpenAI-GPT-5.1 achieves only 62.07\% accuracy, with model performance displaying clear gradient distributions across both reasoning models (16.04\%--62.07\%) and chat models (9.71\%--50.40\%).
\benchmark achieves robust LLM evaluation through contamination-resistant dynamic generation and comprehensive multi-statement assessment.
}

\date{\today}
\correspondence{\email{\{duyantao, huang.wenhao\}@bytedance.com}, \email{jjzhang@nlpr.ia.ac.cn}}

\checkdata[Project Page]{\url{https://encyclo-k.github.io}}

\begin{document}
\maketitle


\section{Introduction}

As LLMs advance rapidly, benchmarks for language understanding evolve accordingly. Early benchmarks such as GLUE~\citep{DBLP:conf/iclr/WangSMHLB19} and SuperGLUE~\citep{DBLP:conf/nips/WangPNSMHLB19} play a pivotal role in advancing fundamental language understanding tasks. To keep pace with improving model capabilities, subsequent benchmarks including MMLU~\citep{hendrycks2020measuring}, MMLU-Pro~\citep{wang2024mmlupro}, TriviaQA~\citep{joshi2017triviaqa0}, and HotpotQA~\citep{yang2018hotpotqa0} raise the bar by broadening knowledge coverage and increasing task complexity. More recently, GPQA~\citep{rein2023gpqa}, SuperGPQA~\citep{pteam2025supergpqascalingllmevaluation}, and HLE~\citep{phan2025humanity0s} push the frontier further by demanding expert-level knowledge comprehension and deep reasoning.

Despite this progress, these benchmarks predominantly collect questions that already exist in textbooks or on websites. This paradigm of curating benchmarks at the question level suffers from three key limitations. First, questions or their variants may already appear in model training corpora. Even if a benchmark is initially uncontaminated, it risks exposure in future training, potentially distorting evaluation results. Second, each question typically targets a single knowledge concept, making it difficult to assess comprehensive understanding across multiple knowledge statements. Third, curating high-quality questions requires domain experts to design, annotate, and review them, incurring substantial cost and time.

To address these limitations, we introduce \benchmark (a dynamic evaluation of encyclopedic knowledge), which rethinks benchmark construction from the ground up. Our key insight is that knowledge statements, not questions, can serve as the unit of curation, and questions can then be constructed from them.

\begin{figure*}[htbp]
    \begin{center}
    \includegraphics[width=\linewidth]{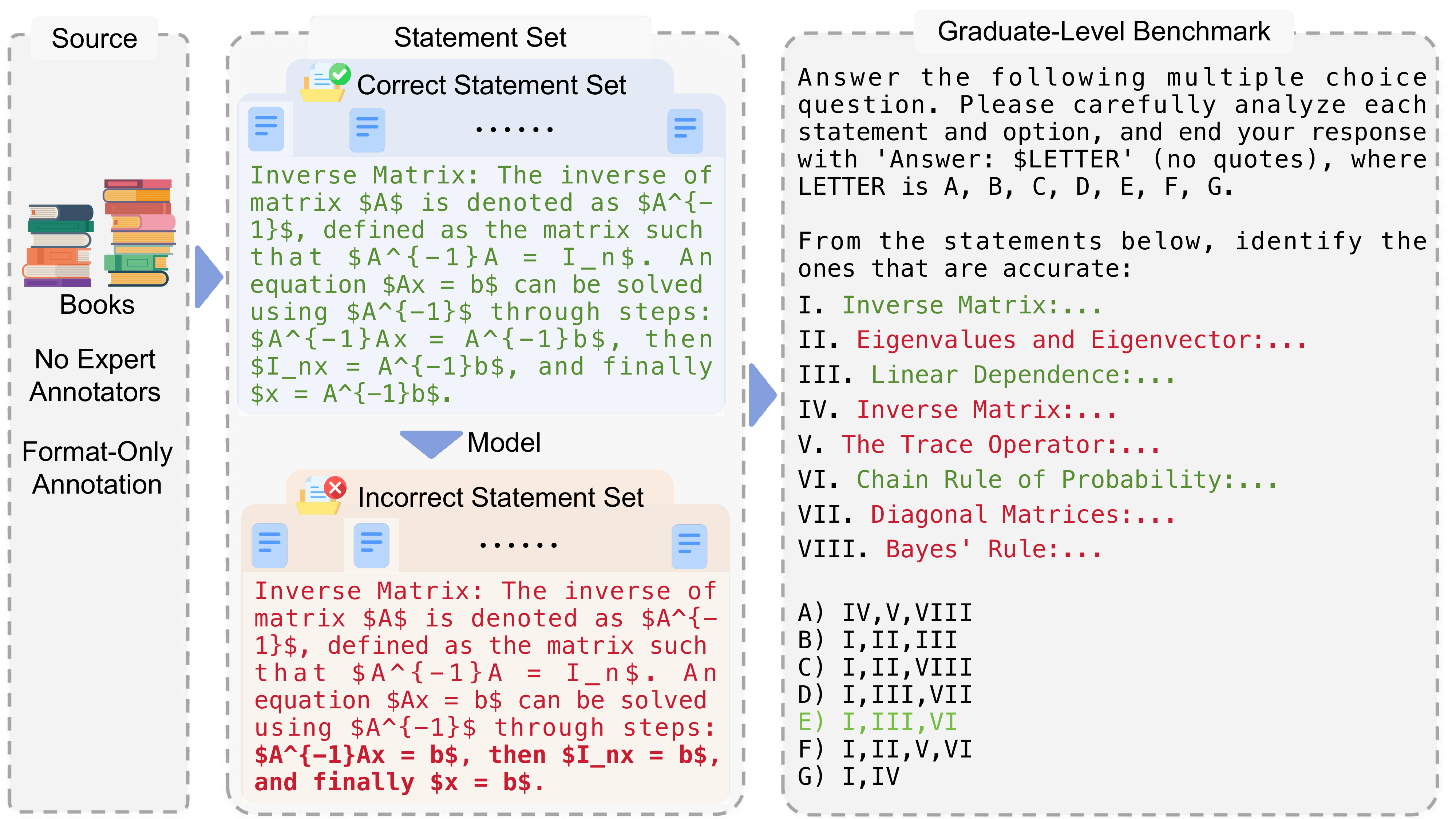}
    \end{center}
    \caption{\textbf{Overview of \benchmark:} from statements to dynamically assembled evaluation questions.}
    \label{Fig: pipeline}
\end{figure*}

Concretely, as illustrated in~\autoref{Fig: pipeline}, we extract standalone knowledge statements from authoritative textbooks and compose them into evaluation questions through random sampling at test time. This simple pipeline directly addresses all three limitations. For contamination resistance, we randomly sample and combine statements into questions at each test time. Even if individual statements appear in training data, their compositions form a combinatorial space too vast to memorize. For comprehensive assessment, each question aggregates 8–10 statements and asks models to identify which are correct. This requires joint understanding across diverse statements, going beyond what single-concept questions can probe. For cost reduction, correct statements are extracted directly from textbooks while incorrect ones are generated by reasoning models. Annotators only verify that each statement is self-contained and correctly formatted, requiring no expertise.


Extensive experiments on over 50 LLMs validate the effectiveness of \benchmark. Regarding \textbf{dynamic evaluation}, model rankings remain stable across five question sets generated with different random seeds, confirming that our combinatorial design resists memorization and enables reliable periodic dataset refresh. Regarding \textbf{multi-statement comprehension}, compared to single-concept questions in traditional benchmarks, models consistently exhibit over 20 percentage points lower accuracy on our multi-statement comprehensive understanding tasks, demonstrating the substantial challenges introduced by joint comprehension of multiple knowledge statements. 

In summary, this paper makes the following contributions:
\begin{itemize}
    \item We rethink benchmark construction by treating individual knowledge statements, rather than questions, as the atomic unit of curation. This shift enables diverse sourcing from authoritative materials, eliminates the need for expert annotation, and substantially reduces construction costs.
    \item We compose statements into questions via random sampling, yielding two key properties: (1) \textbf{Dynamic}—the combinatorial space is too vast to memorize, with experiments across multiple random seeds confirming stable model rankings; (2) \textbf{Multi-statement}—each question requires joint comprehension of multiple statements, posing greater challenges than single-concept questions.
    \item We conduct comprehensive evaluation across 50+ LLMs, demonstrating that multi-statement assessment poses substantial challenges even for state-of-the-art models, while effectively differentiating capabilities across model scales. Additional analyses reveal systematic performance degradation with increasing option counts and confirm the absence of position bias.
\end{itemize}

\section{Related Work}
\subsection{Large Language Models}
LLMs have advanced rapidly in recent years, giving rise to numerous representative models. Among closed-source commercial models, the GPT~\citep{gpt4}, Claude~\citep{Claude}, Gemini~\citep{comanici2025gemini}, and Grok~\citep{grok} achieve continuous breakthroughs in instruction-following and complex reasoning capabilities through large-scale high-quality pre-training, post-training alignment, and inference-time scaling techniques, and demonstrate remarkable levels of general intelligence. In the open-source ecosystem, models such as Llama~\citep{llama3modelcard}, Qwen~\citep{yang2025qwen3}, DeepSeek~\citep{deepseek-ai2024deepseek0v3}, Yi~\citep{ai2024yi0}, and Gemma~\citep{team2025gemma} lower the barriers to research by releasing model weights, fostering widespread adoption and innovation across both academia and industry. Furthermore, fully open-source models represented by OLMo~\citep{olmo20242} and MAP-Neo~\citep{zhang2024mapneo} release not only model weights but also training data, code, and intermediate checkpoints, providing valuable transparent resources for gaining deeper insights into LLM training mechanisms. 
As LLMs continue to advance, robust evaluation becomes increasingly critical. We introduce \benchmark, which features dynamic evaluation and multi-expert-statement comprehensive understanding, enabling deeper knowledge integration assessment while achieving contamination resistance.

\subsection{LLMs Evaluation Benchmarks}

Existing benchmarks evaluate large language models across multiple dimensions, covering language understanding~\citep{DBLP:conf/iclr/WangSMHLB19, DBLP:conf/nips/WangPNSMHLB19, hendrycks2020measuring, wang2024mmlupro, rein2023gpqa, pteam2025supergpqascalingllmevaluation}, reasoning~\citep{suzgun2022challenging, joshi2017triviaqa0, yang2018hotpotqa0}, coding~\citep{chen2021codex, austin2021program, jimenez2024swebench, jain2024livecodebench}, mathematics~\citep{cobbe2021gsm8k, balunovic2025matharena, an2025amo0bench0}, long context~\citep{bai2023longbench0, zhang-etal-2024-bench, DBLP:conf/iclr/WuHHL25}, multilingual capabilities ~\citep{hu2020xtreme, singh2024global, xuan2025mmluproxmultilingualbenchmarkadvanced}, as well as dialogue and instruction-following~\citep{zhou2023instruction0following, zheng2023judging, lin2024wildbench0, jiang2023followbench0}. Additionally, some benchmarks focus on dynamic evaluation, aiming to assess models' adaptation to time-varying knowledge and reduce data contamination risks~\citep{tang-etal-2025-evowiki, jain2024livecodebench}.

To ensure a robust evaluation of model generalizability and alleviate question leakage, we design two distinct mechanisms.
First, \benchmark supports  supports \textbf{dynamic evaluation}: statements within questions can be substituted, making the question content itself dynamically variable and enabling continuous generation of novel test samples to effectively address data contamination.
Second, each question integrates \textbf{multiple knowledge statements}, emphasizing comprehensive understanding while ensuring that individual questions better represent the overall knowledge structure of a discipline. 

\section{\benchmark Dataset}
\label{sec:dataset}

\subsection{Dataset Construction Pipeline}

We introduce a scalable pipeline for constructing expert-level benchmarks without requiring domain expert annotation. As illustrated in~\autoref{Fig: pipeline}, the pipeline consists of three stages: correct statement collection, incorrect statement generation, and dynamic question assembly. The following subsections detail each stage.

\subsubsection{Correct Statement Collection and Filtering}
\label{subsubsec:collection}

We carefully select 62 representative university-level and graduate-level textbooks covering 43 subfields across 11 disciplines. Knowledge statements are manually extracted from these textbooks through screenshot capture. Irrelevant content is masked with mosaics, and doubao-1.5-vision-pro~\cite{volcengine2025doubao} is employed to extract the knowledge statements in the images and generate corresponding knowledge statement titles. Unlike traditional methods requiring domain experts to collect each question and ensure correctness, our approach only requires format validation to obtain correct expert-level statements. The extraction prompt and the collected correct statement instances are provided in \Cref{appendix:pipeline_stage1}.

Based on identified common error patterns (see \Cref{appendix:quality_control} for quality control details), we establish the following specifications that correct statements must satisfy:

\begin{tcolorbox}[colframe=boxcolor,colback=white,boxrule=0.5mm,arc=0mm]
{\Large \textcolor{bulbcolor}{\raisebox{-0.15em}{\faGavel}}\hspace{0.5em}\textbf{Specifications for Correct Statements:}}
\vspace{1mm}
\begin{itemize}
    \item \textbf{Independence and Completeness:} Statements must be independent knowledge units, without referencing external chapters, paragraphs, or external formulas, chemical reaction equations, etc., and must provide complete descriptions of the research subject.
    \item \textbf{Semantic Clarity:} Avoid using ambiguous expressions such as "our country" or "this region," ensuring that statements retain explicit meaning when detached from context.
    \item \textbf{Information Conciseness:} Remove redundant elements, such as formula numbers used for cross-referencing in books; Formulas should directly use expressions without additional reference numbers.
\end{itemize}
\end{tcolorbox}


\subsubsection{Generating Incorrect Statements}
\label{subsubsec:generation}

We leverage DeepSeek-R1~\citep{deepseek-ai2025deepseek0r10} to transform correct statements into incorrect counterparts. We specifically enforce that these fabricated statements maintain a professional tone, logical coherence, and high deceptiveness, while also generating the underlying reasoning for each error. The error taxonomy encompasses concept substitution, causal inversion, detail falsification, temporal dislocation, logical paradox, and scope modification. The specific prompts and illustrative examples are detailed in \Cref{appendix:pipeline_stage2}. We design the generation prompt with the following considerations:

\begin{tcolorbox}[colframe=boxcolor,colback=white,boxrule=0.5mm,arc=0mm]
{\Large \textcolor{bulbcolor}{\raisebox{-0.15em}{\faLightbulb}}\hspace{0.5em}\textbf{Prompt Design Considerations:}}
\vspace{1mm}
\begin{itemize}
    \item \textbf{Definitive Error:} Generated incorrect statements contain explicit errors without controversy.
    \item \textbf{Professionalism:} Use accurate academic terminology and standardized expressions, maintaining a professional standard comparable to correct statements.
    \item \textbf{Logical Self-consistency:} Complete logical structure without obvious internal contradictions.
    \item \textbf{Deceptiveness:} Require deep thinking to identify errors.
    \item \textbf{Cognitive Value:} Error points should reveal typical cognitive misconceptions or comprehension difficulties in the field, rather than peripheral information.
\end{itemize}
\end{tcolorbox}

\subsubsection{Dynamic Question Generation Mechanism}
\label{subsubsec:mechanism}
We combine correct and incorrect statements into a candidate set, from which statements are sampled to populate question templates and generate corresponding options. The question construction process is programmatically controlled, including the assignment of option letters and the generation of option content. This automated pipeline introduces no risk of additional errors. Therefore, ensuring the accuracy of correct statements and the inaccuracy of incorrect statements is sufficient to guarantee the quality of the final questions. Question templates and examples are provided in~\Cref{appendix:pipeline_stage3}. The question generation process supports multi-dimensional parameter configuration: the statement count controls the total number of statements per question, the option count determines the number of candidate options per question, and the combination count specifies the number of statements combined within each option.

\subsection{Dataset Overview}
\subsubsection{Statement Statistics}

\begin{table}[!ht]
    \centering
\caption{Statistics of Statement. Tokens are calculated with Tiktoken using \texttt{\href{https://huggingface.co/BEE-spoke-data/cl100k_base}{cl100k\_base}} encoding.}
\small
\resizebox{\textwidth}{!}{%
\begin{tabular}{@{}lcccccccccc@{}}
\toprule
\multirow{3}{*}{Discipline} & \multicolumn{10}{c}{Statements Statistics} \\
\cmidrule(lr){2-11}
 & \multicolumn{5}{c}{\#Nums} & \multicolumn{3}{c}{\#Tokens} & \multicolumn{2}{c}{\#Coverage} \\
\cmidrule(lr){2-6}\cmidrule(lr){7-9}\cmidrule(lr){10-11}
& Total & En & Zh & Correct & Incorrect & Max & Min & Avg & Fields & SubF \\
\midrule
Science & 10606 & 7317 & 3289 & 5309 & 5297 & 4052 & 8 & 316.56 & 8 & 14 \\
Engineering & 7486 & 4972 & 2514 & 3744 & 3742 & 3684 & 10 & 303.03 & 18 & 23 \\
Literature & 7053 & 1877 & 5176 & 3529 & 3524 & 4768 & 18 & 456.15 & 5 & 8 \\
Medicine & 6300 & 5420 & 880 & 3152 & 3148 & 5366 & 16 & 295.06 & 4 & 5 \\
History & 5892 & 4392 & 1500 & 2948 & 2944 & 3896 & 37 & 311.24 & 1 & 2 \\
Law & 1620 & 1620 & / & 810 & 810 & 2117 & 16 & 292.17 & 1 & 1 \\
Economics & 1052 & 354 & 698 & 526 & 526 & 2133 & 39 & 318.28 & 1 & 2 \\
Agriculture & 955 & 955 & / & 478 & 477 & 1119 & 31 & 233.48 & 1 & 1 \\
Education & 929 & 929 & / & 465 & 464 & 1319 & 40 & 257.88 & 2 & 2 \\
Management & 615 & 158 & 457 & 308 & 307 & 2907 & 73 & 499.34 & 2 & 2 \\
Philosophy & 511 & 54 & 457 & 256 & 255 & 5307 & 56 & 464.15 & 1 & 2 \\
\midrule
\textbf{Overall} & \textbf{43019} & \textbf{28045} & \textbf{14974} & \textbf{21525} & \textbf{21494} & \textbf{5366} & \textbf{8} & \textbf{333.59} & \textbf{44} & \textbf{62} \\
\bottomrule
\end{tabular}
}
\label{tab:statement_stats}
\end{table}
As shown in~\autoref{tab:statement_stats}, we manually collect 21,525 qualified statements from textbooks spanning 11 disciplines, 44 fields, and 62 subfields. Based on these statements, we generated 21,494 incorrect statements, constituting a collection of 43,019 statements. The distribution of statement counts across disciplines exhibits significant variation: STEM disciplines contain 18,092 statements, accounting for 42\% of the entire statement collection, while disciplines with fewer statements include Education, Management, and Philosophy. This imbalanced distribution may be relate to the uneven exposure levels and accessible information sources of disciplines in the real world.

We employ the \texttt{\href{https://github.com/openai/tiktoken}{Tiktoken}} toolkit to measure statement token length. The average length across the entire collection is 333.59 tokens, with disciplinary averages ranging from 233.48 (Agriculture) to 499.34 (Management). Individual statement lengths span from 8 to 5,366 tokens, indicating that while certain variations exist across disciplines, the overall lengths fall within a reasonable range for knowledge point descriptions.

\subsubsection{Question Statistics}

\begin{table}[!ht]
    \centering
\caption{Statistics of \benchmark Question. Tokens are calculated with Tiktoken using \texttt{\href{https://huggingface.co/BEE-spoke-data/cl100k_base}{cl100k\_base}} encoding.}
\small
\resizebox{\textwidth}{!}{%
\begin{tabular}{@{}lcccccccccccccccc@{}}
\toprule
\multirow{3}{*}{Discipline} & \multicolumn{15}{c}{Questions Statistics} \\
\cmidrule(lr){2-16}
& \multicolumn{3}{c}{\#Nums} & \multicolumn{3}{c}{\#Tokens} & \multicolumn{3}{c}{\#Statements} & \multicolumn{3}{c}{\#Options} & \multicolumn{3}{c}{\#Combinations} \\
\cmidrule(lr){2-4}\cmidrule(lr){5-7}\cmidrule(lr){8-10}\cmidrule(lr){11-13}\cmidrule(lr){14-16}
& Total & En & Zh & Max & Min & Avg & Max & Min & Avg & Max & Min & Avg & Max & Min & Avg \\
\midrule
Science & 1242 & 858 & 384 & 13223 & 715 & 2992.46 & 10 & 8 & 8.99 & 8 & 4 & 5.98 & 4 & 2 & 3.00 \\
Engineering & 881 & 583 & 298 & 12206 & 532 & 2840.50 & 10 & 8 & 9.01 & 8 & 4 & 6.06 & 4 & 2 & 3.02 \\
Literature & 825 & 221 & 604 & 16474 & 992 & 4169.00 & 10 & 8 & 8.99 & 8 & 4 & 5.96 & 4 & 2 & 3.01 \\
Medicine & 736 & 634 & 102 & 16321 & 797 & 2780.22 & 10 & 8 & 9.01 & 8 & 4 & 6.00 & 4 & 2 & 2.99 \\
History & 686 & 512 & 174 & 10557 & 1312 & 2883.96 & 10 & 8 & 8.96 & 8 & 4 & 6.10 & 4 & 2 & 3.03 \\
Law & 189 & 189 & / & 6667 & 1508 & 2854.86 & 10 & 8 & 9.04 & 8 & 4 & 6.05 & 4 & 2 & 2.99 \\
Economics & 124 & 42 & 82 & 5433 & 1582 & 3026.01 & 10 & 8 & 8.98 & 8 & 4 & 6.22 & 4 & 2 & 3.04 \\
Agriculture & 112 & 112 & / & 3850 & 1112 & 2201.90 & 10 & 8 & 8.94 & 8 & 4 & 5.86 & 4 & 2 & 2.99 \\
Education & 109 & 109 & / & 4088 & 1171 & 2394.45 & 10 & 8 & 8.92 & 8 & 4 & 5.61 & 4 & 2 & 3.02 \\
Management & 73 & 19 & 54 & 7349 & 2353 & 4527.04 & 10 & 8 & 8.90 & 8 & 4 & 6.12 & 4 & 2 & 3.04 \\
Philosophy & 61 & 7 & 54 & 11671 & 1960 & 4074.30 & 10 & 8 & 8.74 & 8 & 4 & 6.07 & 4 & 2 & 3.01 \\
\midrule
\textbf{Overall} & \textbf{5038} & \textbf{3286} & \textbf{1752} & \textbf{16474} & \textbf{532} & \textbf{3113.05} & \textbf{10} & \textbf{8} & \textbf{8.98} & \textbf{8} & \textbf{4} & \textbf{6.00} & \textbf{4} & \textbf{2} & \textbf{3.01} \\
\bottomrule
\end{tabular}
}
\label{tab:question_stats}
\end{table}

As shown in~\autoref{tab:question_stats}, our dataset comprises 5,038 questions, including 3,286 English questions and 1,752 Chinese questions. The disciplinary distribution of questions is strictly designed proportional to statement ratios: the Science discipline has the most questions (1,242, accounting for 24.7\%), while Philosophy has the fewest (61, accounting for 1.2\%), ensuring consistency between question-level and knowledge point-level disciplinary distributions. Question lengths exhibit significant variation, with an average of 3,113.05 tokens. Disciplinary ranges span from 2,201.90 (Agriculture) to 4,527.04 (Management), fully reflecting the inherent differences in question complexity and information density across disciplines. Details regarding the division and quantity of fields for each discipline can be found in~\Cref{appendix:question_distribution}.

Benefiting from the programmatic generation process, question structure exhibits highly standardized characteristics: each question contains 8--10 statements (average 8.98), 4--8 options (average 6.00), and 2--4 combination answers (average 3.01). This consistent format design enables question comparability and provides a foundation for standardized evaluation. Overall, the dataset maintains disciplinary proportion consistency while achieving a balance between breadth of knowledge coverage and evaluation difficulty through standardized question structure and diverse content lengths.

\subsubsection{Semantic Visualization}
\begin{figure*}[ht]
\begin{center}
\includegraphics[width=0.47\linewidth]{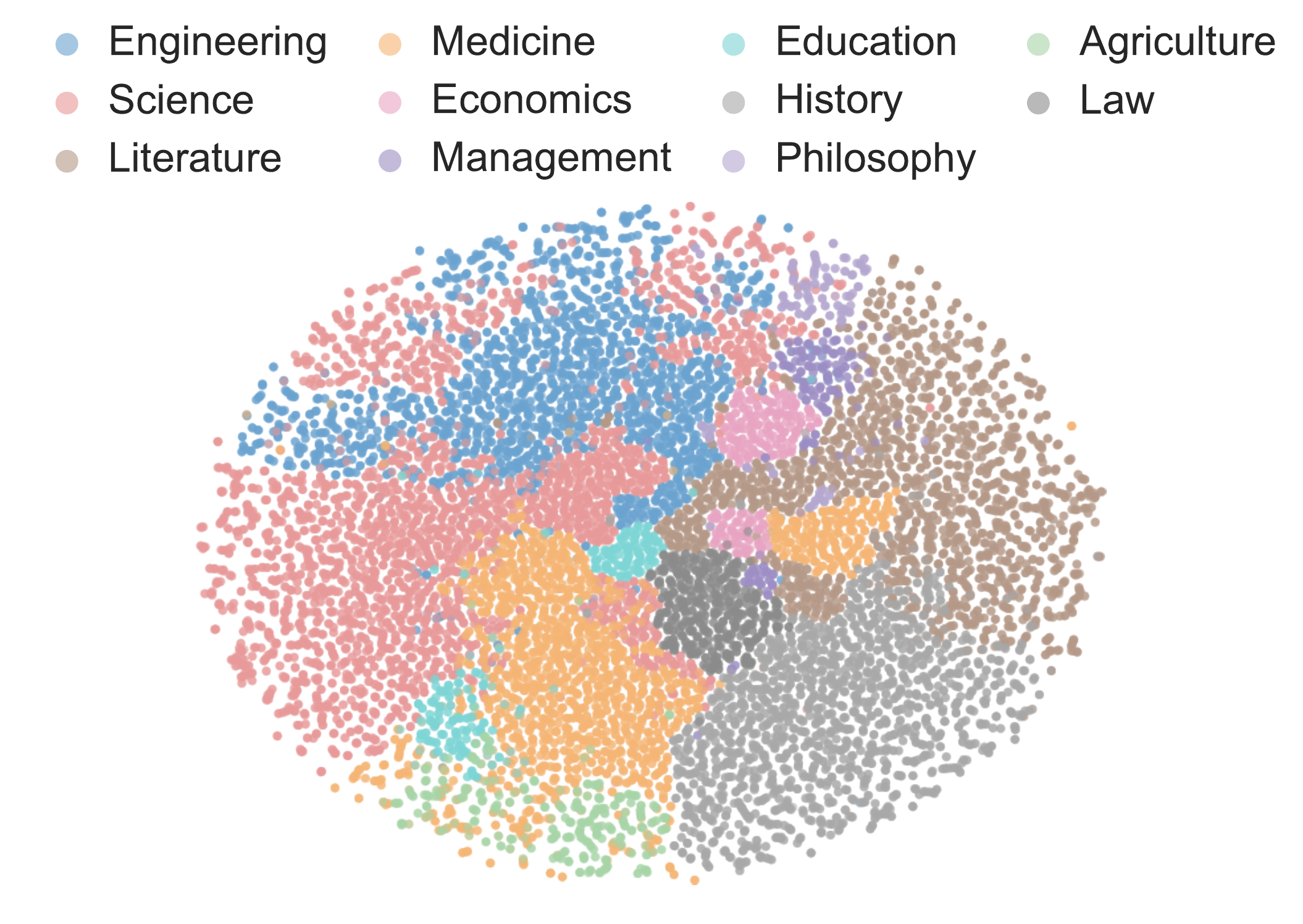}
\hspace{2em}
\includegraphics[width=0.47\linewidth]{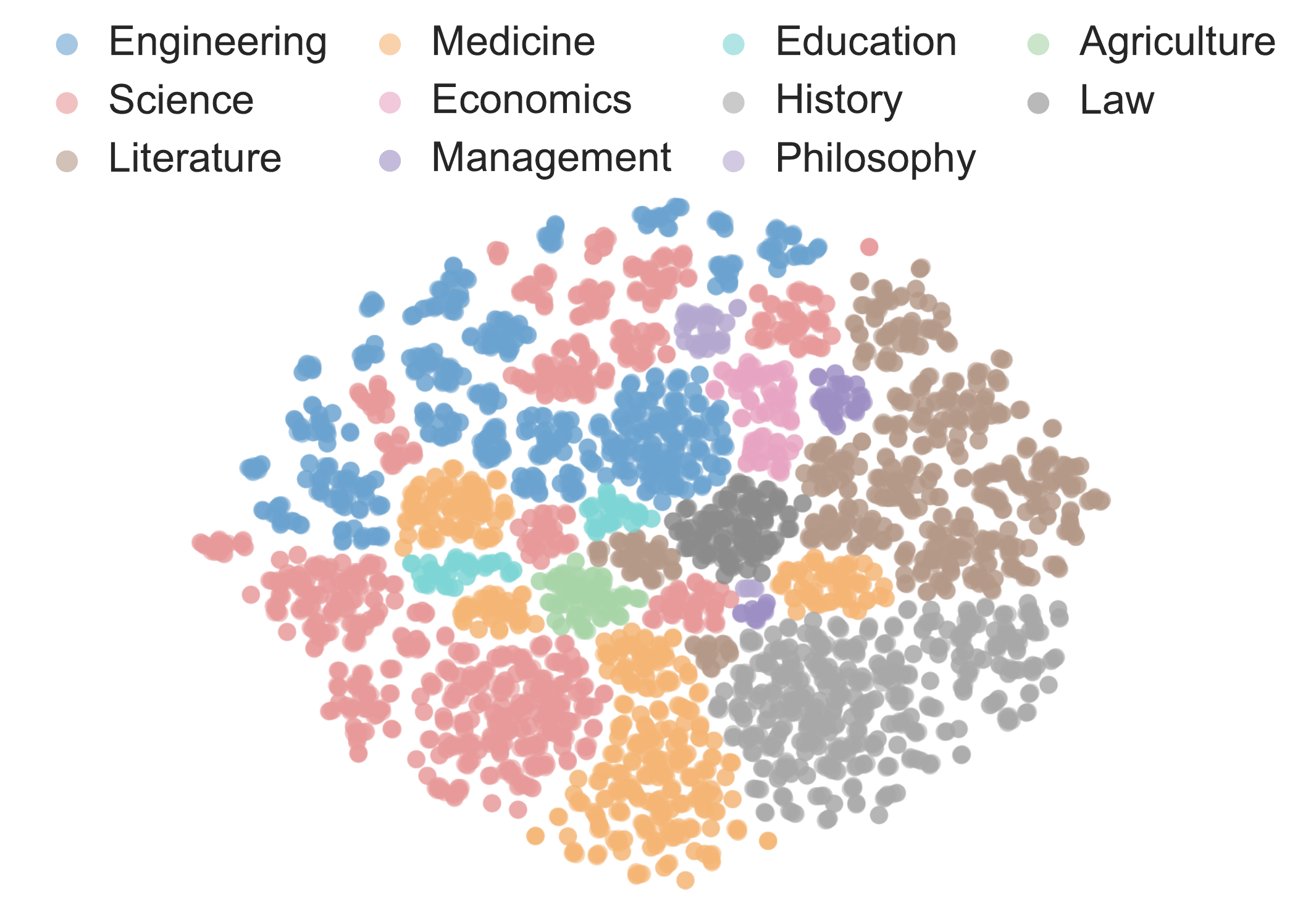}
\end{center}
\caption{\textbf{Comparison of semantic distributions.} knowledge statements exhibit cross-disciplinary overlap (left), while generated questions demonstrate distinct disciplinary clustering (right).}
\label{fig:semantic_viz}
\end{figure*}

As shown in~\autoref{fig:semantic_viz}, we project the collected statements and generated questions separately. In the statement embedding space (\autoref{fig:semantic_viz}, left), knowledge points from different disciplines exhibit a relatively mixed distribution with relatively blurred disciplinary boundaries, reflecting the cross-disciplinary commonality and interconnectedness of basic knowledge units at the semantic level. For example, statements from Science and Engineering disciplines show obvious overlap in the embedding space, manifesting the similarity of STEM disciplines in fundamental concepts and methodologies.

In contrast, the question-level visualization (\autoref{fig:semantic_viz}, right) demonstrates a clearer disciplinary clustering structure. Each discipline forms relatively independent semantic clusters with more distinct disciplinary boundaries. This phenomenon indicates that although individual knowledge points may possess cross-disciplinary attributes, when multiple statements are combined into complete questions, disciplinary characteristics are significantly amplified. The density of domain-specific terminology, reasoning patterns, and knowledge organization methods in questions work together to form distinguishable distribution patterns for different disciplines in high-dimensional semantic space. Notably, humanities and social sciences such as Law and History show clear separation from STEM disciplines in the embedding space, while Science, Engineering, and Medicine form adjacent but distinguishable clustering groups.

This semantic structure evolution from statements to questions validates the effectiveness of our dataset design: by combining multiple knowledge points to construct questions, we successfully aggregate and strengthen disciplinary characteristics. This enables the dataset to maintain fine-grained knowledge granularity while possessing clear disciplinary differentiation, providing a solid foundation for subsequent evaluation of models' cross-disciplinary knowledge understanding capabilities and disciplinary specialization levels.
\section{Experimental Setup}

\subsection{Dynamic Question Generation Setting}
For dynamic question generation, we configure the following parameters. Each question comprises 8 to 10 knowledge statements and provides 4 to 8 candidate options. We construct each option by combining 2 to 4 statements, thereby creating diverse statement combinations as candidate answers. We adopt a discipline-proportional allocation strategy for question generation. Under the standard configuration, we generate 5,038 evaluation questions in total, allocating questions to each discipline by rounding up according to its proportional corpus size. This configuration ensures that the disciplinary distribution of the evaluation set aligns with that of the knowledge base. We format all statements within questions using lowercase Roman numerals (i., ii., iii., ...) and identify options with uppercase letters (A, B, C, ...).

\subsection{Answer Extraction}

To extract answers from model-generated responses, we employ a multi-tier regular expression matching strategy. The system initially applies three pattern groups to match: (1) standard formats (e.g., The answer is: A), (2) simplified formats (e.g., Answer: A), and (3) isolated letters (e.g., A, (B)). Matching first occurs on the last line of the generated text; if unsuccessful, it extends to the full text. When all regex patterns fail to retrieve a valid response, a fallback mechanism directly searches for text segments matching option content. If this still fails to extract a valid answer, the sample is marked as \textit{miss} and counted as incorrect. This ensures consistent and reliable answer extraction across all evaluations.

\section{Results and Analysis}
\subsection{Main Results}
\definecolor{color11}{rgb}{1, 0.8, 0.8}  
\definecolor{color12}{rgb}{1, 0.9, 0.9}  
\definecolor{color21}{RGB}{255, 224, 127}  
\definecolor{color22}{RGB}{255, 239, 179}  
\definecolor{color31}{RGB}{198, 230, 195}  
\definecolor{color32}{RGB}{224, 239, 225}  
{
\begin{table}[htbp]
\small
\centering
\captionsetup{font=footnotesize}
\caption{\textbf{Performance Overview on \benchmark. } 
The table presents LLMs' performance on different disciplines. The highest score in each column is indicated with a \boxed{box}; the second-best score is in \textbf{bold}, and the third-best score is \underline{underlined}.}
\label{performance2_all}
\setlength{\tabcolsep}{2pt} 
\resizebox{\textwidth}{!}{%
\begin{tabular}{lcccccccccccc}
\toprule
\textbf{Model} & \textbf{Avg.} & \textbf{Sci.} & \textbf{Eng.} & \textbf{Lit.} & \textbf{Med.} & \textbf{Hist.} & \textbf{Law} & \textbf{Econ.} & \textbf{Agr.} & \textbf{Edu.} & \textbf{Mgt.} & \textbf{Phil.}\\
\midrule
\rowcolor{color11}
\multicolumn{13}{c}{\textbf{\textit{Reasoning Models}}}\\
\midrule
\rowcolor{color12}
OpenAI-GPT-5.1-high & \boxed{62.07} & \boxed{64.09} & \boxed{66.74} & \underline{55.15} & \textbf{65.08} & \textbf{51.17} & \boxed{70.37} & \textbf{70.16} & \boxed{56.25} & \textbf{73.39} & \boxed{68.49} & \boxed{73.77} \\
\rowcolor{color12}
Gemini-3-Pro-Preview-Exp & \textbf{61.75} & \textbf{62.56} & \underline{63.00} & \boxed{56.97} & \underline{65.08} & \boxed{57.43} & 64.02 & \boxed{74.19} & \underline{51.79} & 69.72 & \textbf{67.12} & 65.57 \\
\rowcolor{color12}
Gemini-2.5-Pro & \underline{58.93} & \underline{61.35} & 58.12 & \textbf{55.64} & 61.96 & \underline{51.17} & \textbf{66.67} & \underline{68.55} & 46.43 & \underline{70.64} & 61.64 & \textbf{72.13} \\
\rowcolor{color12}
OpenAI-o3-high.code & 58.36 & 60.95 & \textbf{63.22} & 49.70 & \boxed{65.35} & 47.52 & 57.67 & 62.90 & \textbf{52.68} & \boxed{75.23} & 56.16 & 65.57 \\
\rowcolor{color12}
OpenAI-GPT-5-high.code & 57.26 & 59.26 & 60.84 & 49.21 & 63.86 & 46.21 & \underline{66.14} & 64.52 & 50.89 & 62.39 & \underline{64.38} & \underline{70.49} \\
\rowcolor{color12}
Qwen3-235B-A22B-Thinking-2507 & 51.77 & 55.88 & 54.71 & 47.76 & 54.48 & 45.19 & 43.39 & 55.65 & 33.04 & 61.47 & 47.95 & 60.66 \\
\rowcolor{color12}
Doubao-1.6-thinking.foreval & 50.97 & 51.93 & 51.87 & 47.03 & 55.43 & 47.08 & 49.21 & 53.23 & 35.71 & 65.14 & 54.79 & 60.66 \\
\rowcolor{color12}
DeepSeek-V3.1-Terminus & 49.78 & 51.61 & 49.60 & 47.64 & 52.17 & 45.04 & 52.38 & 57.26 & 36.61 & 55.05 & 42.47 & 68.85 \\
\rowcolor{color12}
DeepSeek-R1 & 48.99 & 51.21 & 49.60 & 45.33 & 52.99 & 44.75 & 50.26 & 56.45 & 33.93 & 54.13 & 45.21 & 47.54 \\
\rowcolor{color12}
Claude-4-Sonnet-thinking-azure & 48.57 & 50.56 & 51.53 & 45.94 & 50.41 & 40.09 & 49.74 & 57.26 & 36.61 & 67.89 & 36.99 & 54.10 \\
\rowcolor{color12}
OpenAI-o1-1217.high.code & 46.80 & 48.71 & 43.47 & 40.48 & 51.90 & 43.59 & 50.26 & 57.26 & 50.00 & 58.72 & 49.32 & 54.10 \\
\rowcolor{color12}
Qwen3-235B-A22B & 43.25 & 41.38 & 44.38 & 43.88 & 48.91 & 37.90 & 40.21 & 50.00 & 28.57 & 49.54 & 45.21 & 57.38 \\
\rowcolor{color12}
QwQ-32B & 40.97 & 41.79 & 44.49 & 37.58 & 44.84 & 34.84 & 35.45 & 48.39 & 20.54 & 60.55 & 34.25 & 54.10 \\
\rowcolor{color12}
Qwen3-30B-A3B-Thinking-2507 & 40.57 & 43.56 & 43.25 & 37.45 & 44.02 & 34.99 & 36.51 & 40.32 & 25.00 & 43.12 & 31.51 & 52.46 \\
\rowcolor{color12}
Qwen3-32B & 37.28 & 37.04 & 40.41 & 35.76 & 40.62 & 32.65 & 31.22 & 38.71 & 29.46 & 43.12 & 38.36 & 47.54 \\
\rowcolor{color12}
Qwen3-4B-Thinking-2507 & 33.45 & 38.08 & 35.98 & 29.94 & 33.29 & 29.15 & 21.69 & 37.10 & 22.32 & 41.28 & 27.40 & 42.62 \\
\rowcolor{color12}
Qwen3-30B-A3B & 31.56 & 32.13 & 33.48 & 32.48 & 33.56 & 28.43 & 27.51 & 29.03 & 19.64 & 38.53 & 21.92 & 29.51 \\
\rowcolor{color12}
Qwen3-14B & 30.79 & 32.61 & 34.17 & 28.73 & 32.34 & 26.68 & 28.04 & 18.55 & 25.00 & 37.61 & 28.77 & 34.43 \\
\rowcolor{color12}
Qwen3-8B & 26.78 & 28.26 & 29.06 & 23.52 & 27.31 & 25.22 & 23.81 & 24.19 & 14.29 & 42.20 & 28.77 & 26.23 \\
\rowcolor{color12}
Qwen3-4B & 23.64 & 26.25 & 25.09 & 22.42 & 22.83 & 20.85 & 17.99 & 21.77 & 17.86 & 33.03 & 20.55 & 26.23 \\
\rowcolor{color12}
Qwen3-1.7B & 18.30 & 17.87 & 20.20 & 18.79 & 17.66 & 17.78 & 16.93 & 17.74 & 16.96 & 15.60 & 17.81 & 19.67 \\
\rowcolor{color12}
Qwen3-0.6B & 16.04 & 16.02 & 17.14 & 16.61 & 13.86 & 15.45 & 13.76 & 19.35 & 19.64 & 13.76 & 15.07 & 24.59 \\
\midrule
\rowcolor{color21}
\multicolumn{13}{c}{\textbf{\textit{Chat Models}}}\\
\midrule
\rowcolor{color22}
Qwen3-235B-A22B-Instruct-2507 & \boxed{50.40} & \boxed{53.30} & \boxed{51.42} & \boxed{48.97} & \boxed{52.17} & \boxed{43.59} & \boxed{47.62} & \boxed{56.45} & \boxed{34.82} & \boxed{56.88} & \boxed{43.84} & \boxed{72.13} \\
\rowcolor{color22}
Kimi-K2-Instruct-0905 & \textbf{44.66} & \textbf{47.99} & \underline{43.47} & \underline{44.85} & \textbf{45.79} & \textbf{40.96} & 36.51 & \textbf{50.00} & \underline{30.36} & \underline{49.54} & \textbf{39.73} & \textbf{57.38} \\
\rowcolor{color22}
DeepSeek-V3.1-Terminus-non-thinking & \underline{42.70} & 40.98 & \textbf{43.93} & \textbf{45.58} & \underline{45.79} & \underline{38.19} & \textbf{40.74} & 46.77 & 28.57 & \textbf{53.21} & 27.40 & \underline{57.38} \\
\rowcolor{color22}
Kimi-K2.volc.foreval & 40.53 & \underline{41.55} & 41.32 & 40.97 & 39.81 & 36.88 & \underline{40.21} & \underline{49.19} & 26.79 & 44.04 & \underline{39.73} & 55.74 \\
\rowcolor{color22}
Qwen3-30B-A3B-Instruct-2507 & 36.72 & 38.97 & 38.14 & 35.03 & 37.50 & 32.51 & 35.45 & 33.06 & 25.00 & 43.12 & 39.73 & 49.18 \\
\rowcolor{color22}
GLM-4.6 & 36.52 & 33.98 & 37.57 & 36.61 & 39.67 & 33.53 & 39.15 & 40.32 & \textbf{31.25} & 47.71 & 32.88 & 45.90 \\
\rowcolor{color22}
DeepSeek-V3-0324 & 30.85 & 28.58 & 31.78 & 32.24 & 32.34 & 29.88 & 32.80 & 29.84 & 24.11 & 38.53 & 27.40 & 36.07 \\
\rowcolor{color22}
Qwen3-235B-A22B-non-thinking & 28.70 & 26.57 & 31.44 & 31.39 & 28.40 & 26.82 & 26.46 & 29.03 & 13.39 & 37.61 & 31.51 & 36.07 \\
\rowcolor{color22}
Qwen3-4B-Instruct-2507 & 28.60 & 29.79 & 31.44 & 26.30 & 29.08 & 24.93 & 25.93 & 31.45 & 16.96 & 33.94 & 31.51 & 40.98 \\
\rowcolor{color22}
Qwen3-32B-no-thinking & 27.81 & 26.33 & 27.13 & 28.48 & 30.03 & 29.01 & 26.46 & 23.39 & 21.43 & 36.70 & 23.29 & 32.79 \\
\rowcolor{color22}
Qwen3-30B-A3B-non-thinking & 25.59 & 25.76 & 27.92 & 24.61 & 26.09 & 24.20 & 21.16 & 23.39 & 18.75 & 34.86 & 20.55 & 31.15 \\
\rowcolor{color22}
Qwen2.5-32B-Instruct & 25.31 & 21.10 & 27.81 & 27.64 & 25.14 & 25.22 & 27.51 & 22.58 & 22.32 & 37.61 & 20.55 & 34.43 \\
\rowcolor{color22}
Qwen2.5-72B-Instruct & 25.03 & 22.95 & 26.11 & 26.55 & 25.41 & 24.78 & 25.40 & 23.39 & 19.64 & 35.78 & 24.66 & 22.95 \\
\rowcolor{color22}
DeepSeek-V3 & 22.93 & 23.19 & 24.97 & 21.58 & 23.10 & 22.16 & 21.16 & 23.39 & 16.07 & 31.19 & 16.44 & 22.95 \\
\rowcolor{color22}
Llama-3.3-70B-Instruct & 20.07 & 19.00 & 23.95 & 17.94 & 20.24 & 17.20 & 24.87 & 16.13 & 25.00 & 22.94 & 20.55 & 22.95 \\
\rowcolor{color22}
Qwen2.5-14B-Instruct & 20.01 & 18.60 & 22.36 & 17.94 & 21.60 & 21.72 & 21.16 & 12.10 & 17.86 & 24.77 & 9.59 & 24.59 \\
\rowcolor{color22}
Qwen3-8B-no-thinking & 19.61 & 18.68 & 20.09 & 22.42 & 20.24 & 19.68 & 17.46 & 13.71 & 12.50 & 17.43 & 21.92 & 18.03 \\
\rowcolor{color22}
Llama-3.1-70B-Instruct & 19.37 & 19.24 & 21.00 & 19.76 & 18.34 & 16.47 & 19.58 & 19.35 & 18.75 & 31.19 & 15.07 & 22.95 \\
\rowcolor{color22}
Qwen2.5-7B-Instruct & 19.02 & 19.57 & 18.62 & 20.61 & 20.52 & 16.91 & 13.76 & 20.97 & 16.96 & 15.60 & 15.07 & 24.59 \\
\rowcolor{color22}
Qwen3-4B-no-thinking & 18.18 & 18.68 & 17.37 & 18.30 & 19.16 & 18.08 & 16.93 & 12.90 & 12.50 & 22.94 & 17.81 & 24.59 \\
\rowcolor{color22}
Qwen3-14B-no-thinking & 17.98 & 16.43 & 18.73 & 19.27 & 19.43 & 18.66 & 15.87 & 8.87 & 15.18 & 19.27 & 19.18 & 22.95 \\
\rowcolor{color22}
Qwen2.5-3B-Instruct & 17.75 & 16.75 & 17.48 & 19.39 & 18.07 & 19.83 & 14.81 & 13.71 & 13.39 & 18.35 & 19.18 & 14.75 \\
\rowcolor{color22}
Yi-1.5-34B-Chat & 17.65 & 14.57 & 17.37 & 19.76 & 19.02 & 19.68 & 20.63 & 12.10 & 15.18 & 20.18 & 20.55 & 14.75 \\
\rowcolor{color22}
Yi-1.5-9B-Chat & 17.43 & 16.83 & 18.16 & 17.82 & 18.89 & 17.20 & 14.81 & 16.94 & 16.07 & 15.60 & 15.07 & 16.39 \\
\rowcolor{color22}
Qwen2.5-1.5B-Instruct & 17.07 & 17.55 & 15.66 & 18.06 & 17.93 & 17.20 & 16.40 & 15.32 & 10.71 & 20.18 & 16.44 & 14.75 \\
\rowcolor{color22}
Qwen3-1.7B-no-thinking & 17.03 & 17.47 & 15.78 & 17.70 & 16.17 & 16.76 & 14.29 & 14.52 & 22.32 & 22.02 & 17.81 & 24.59 \\
\rowcolor{color22}
MAP-Neo-7B-Instruct-v0.1 & 16.81 & 15.46 & 17.82 & 17.33 & 16.85 & 17.20 & 14.81 & 16.94 & 14.29 & 20.18 & 17.81 & 21.31 \\
\rowcolor{color22}
gemma-2-9b-it & 16.55 & 16.91 & 17.03 & 14.91 & 17.12 & 15.31 & 15.34 & 25.00 & 12.50 & 13.76 & 24.66 & 21.31 \\
\rowcolor{color22}
Yi-1.5-6B-Chat & 15.52 & 15.46 & 15.44 & 14.67 & 14.54 & 17.64 & 17.46 & 12.90 & 15.18 & 9.17 & 19.18 & 24.59 \\
\rowcolor{color22}
Llama-3.1-8B-Instruct & 15.38 & 15.78 & 15.78 & 13.33 & 17.26 & 16.76 & 16.40 & 12.90 & 8.93 & 16.51 & 5.48 & 14.75 \\
\rowcolor{color22}
Qwen3-0.6B-no-thinking & 15.32 & 15.78 & 16.23 & 15.52 & 13.86 & 15.60 & 14.81 & 13.71 & 14.29 & 18.35 & 9.59 & 13.11 \\
\rowcolor{color22}
gemma-2-2b-it & 12.94 & 12.16 & 12.60 & 12.36 & 13.32 & 13.70 & 14.29 & 16.94 & 8.93 & 20.18 & 13.70 & 9.84 \\
\rowcolor{color22}
OLMo-2-1124-7B-Instruct & 12.76 & 13.45 & 11.12 & 10.91 & 12.23 & 14.14 & 14.29 & 12.10 & 17.86 & 31.19 & 2.74 & 4.92 \\
\rowcolor{color22}
Qwen2.5-0.5B-Instruct & 11.61 & 11.03 & 12.15 & 8.00 & 13.72 & 12.39 & 12.17 & 12.10 & 15.18 & 18.35 & 13.70 & 6.56 \\
\rowcolor{color22}
Llama-3.2-3B-Instruct & 11.49 & 12.40 & 11.12 & 9.58 & 11.82 & 12.39 & 13.76 & 8.87 & 18.75 & 11.93 & 4.11 & 3.28 \\
\rowcolor{color22}
Llama-3.2-1B-Instruct & 9.71 & 9.02 & 8.51 & 7.03 & 11.68 & 14.72 & 11.64 & 4.03 & 13.39 & 10.09 & 4.11 & 1.64 \\
\bottomrule
\end{tabular}
}
\end{table}
}

\textbf{Overall Performance.} \benchmark poses a significant challenge and demonstrates strong discriminative power for current large language models. As shown in~\autoref{performance2_all}, even the most powerful reasoning model, OpenAI-GPT-5.1-high, achieves an average score of only 62.07\%. The best-performing chat model, Qwen3-235B-A22B-Instruct-2507, attains merely 50.40\% on average. These results indicate that \benchmark poses a substantial challenge to existing LLMs and remains far from being \textit{solved}.
Meanwhile, the benchmark exhibits excellent discriminative capability. Model performance displays a clear gradient distribution: reasoning models span from 16.04\% to 62.07\%, and chat models range from 9.71\% to 50.40\%. This broad score distribution enables \benchmark to effectively differentiate models across various capability levels, providing fine-grained reference for model evaluation.



\textbf{Discipline-wise Analysis.} Model performance varies substantially across the 11 disciplines. Education and Philosophy emerge as relatively easier domains, with top models exceeding 70\% (e.g., OpenAI-o3-high achieves 75.23\% in Education). In contrast, Agriculture proves most challenging—even the best-performing OpenAI-GPT-5.1-high reaches only 56.25\%, possibly reflecting the scarcity of agricultural expertise in training corpora or the domain's reliance on region-specific practical knowledge. Among top performers, OpenAI-GPT-5.1-high and Gemini-3-Pro-Preview-Exp exhibit complementary strengths: the former excels in STEM and Law, while the latter leads in humanities and Economics.

\textbf{Gap Between Open-weight and Fully Open-source Models.} Fully open-source models (with publicly available training code and data), such as MAP-Neo-7B-Instruct-v0.1 (16.81\%) and OLMo-2-1124-7B-Instruct (12.76\%), lag considerably behind open-weight models of comparable scale. Using the 7-8B parameter range as reference, Qwen2.5-7B-Instruct (19.02\%) and Qwen3-8B (26.78\%) substantially outperform the aforementioned fully open-source models. This disparity may stem from differences in training data quality, scale, and training strategies, suggesting that fully open-source models still have considerable room for improvement in professional knowledge evaluation.

\begin{figure*}[htbp]
\begin{center}
\includegraphics[width=0.46\linewidth]{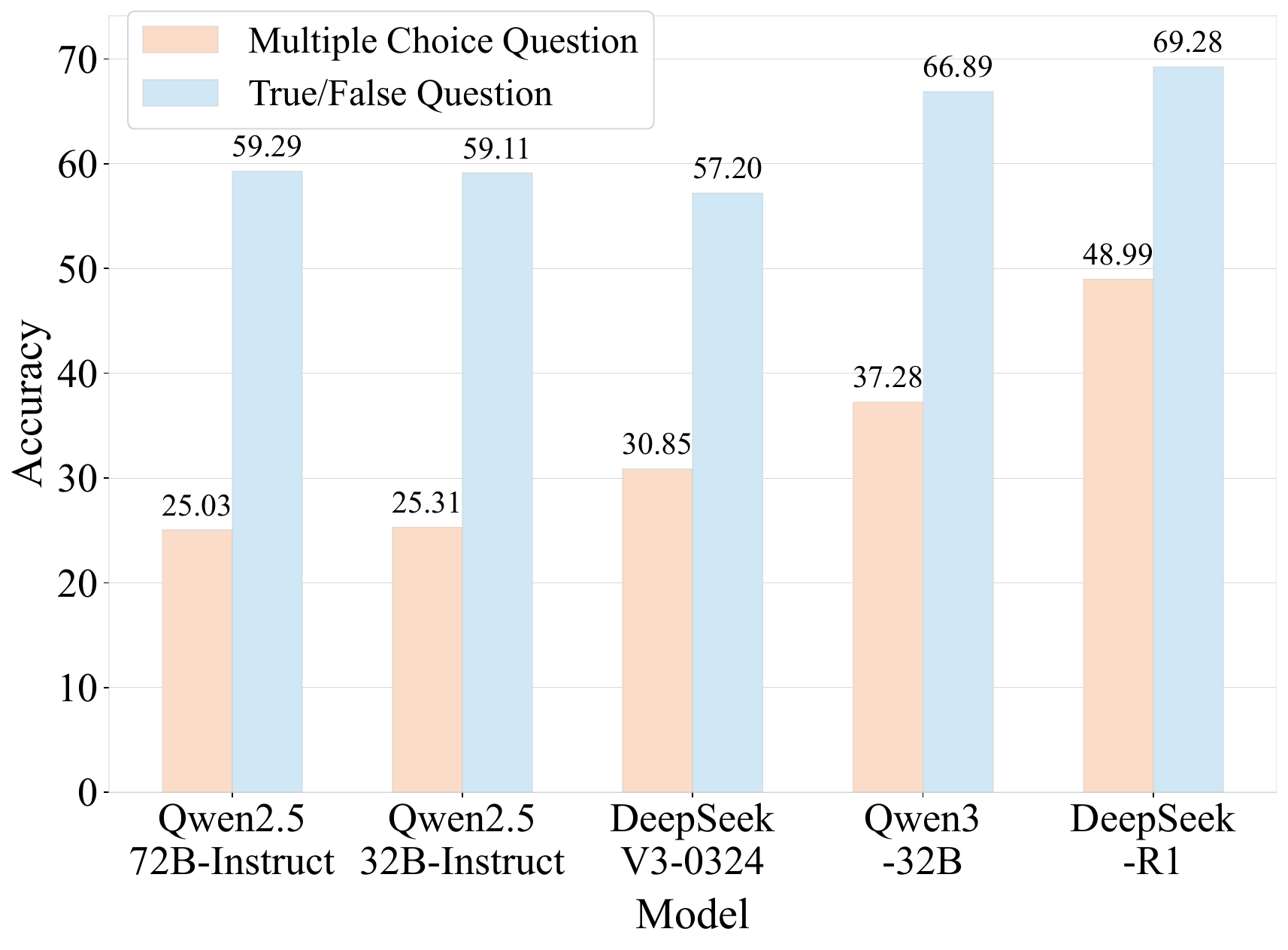}
\hspace{1em}
\includegraphics[width=0.48\linewidth]{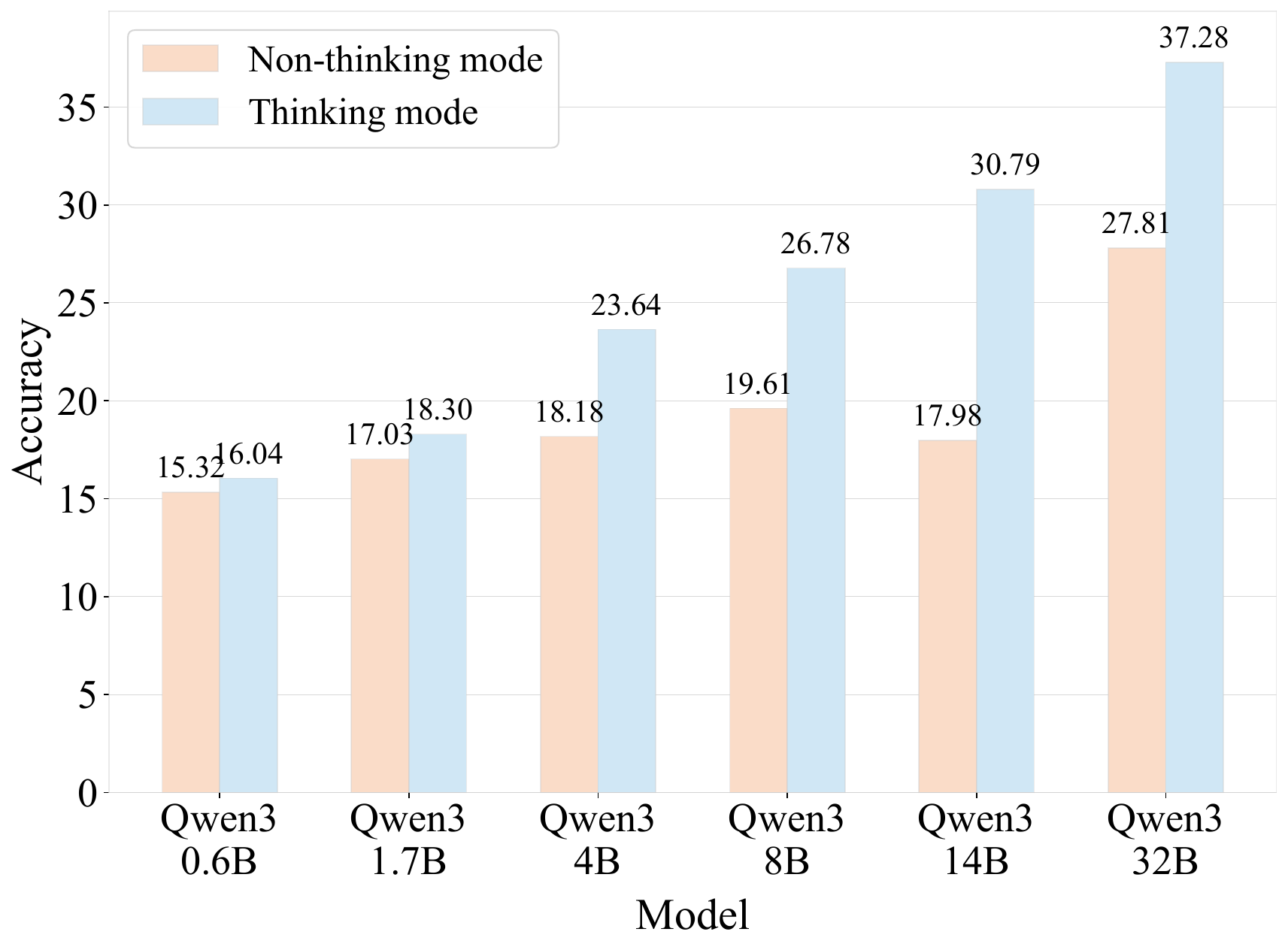}
\end{center}
\caption{\textbf{Left:} Performance comparison between single-statement judgment and multi-statement comprehensive understanding tasks. \textbf{Right:} Accuracy of Qwen3 hybrid models (0.6B--32B) under non-thinking and thinking modes.}
\label{fig:bar_plot}
\end{figure*}

\textbf{Challenge of Multi-Statement Comprehension.} A core design principle of \benchmark is requiring models to jointly comprehend multiple knowledge statements within a single question. To quantify the additional challenge introduced by this multi-statement setting, we compare model performance on single-statement judgment tasks versus multi-statement comprehensive understanding tasks. As shown in~\autoref{fig:bar_plot} (left), models consistently achieve substantially higher accuracy when evaluating individual statements in isolation. For instance, DeepSeek-R1 achieves 69.28\% accuracy on single-statement tasks but drops to 48.99\% on multi-statement questions—a decline of over 20 percentage points. Similar patterns emerge across other models: Qwen3-32B decreases from 66.89\% to 37.28\%, and Qwen2.5-32B-Instruct falls from 59.11\% to 25.31\%. \textbf{This substantial performance gap demonstrates that multi-statement comprehension introduces significant cognitive complexity beyond simple statement-level verification.} The challenge arises from multiple factors: models must simultaneously hold multiple knowledge points in working memory, identify subtle errors across diverse domains, and synthesize judgments into coherent option selections. These findings validate that \benchmark effectively assesses comprehensive understanding capabilities rather than isolated factual recall. Additional experimental settings and detailed analyses are provided in~\Cref{appendix:multi_statement_analysis}.

\textbf{Benefits of Chain-of-Thought Reasoning.} Given that \benchmark requires joint comprehension of multiple knowledge statements, a natural question arises: does explicit reasoning help models synthesize information across statements, or can direct answering suffice? To address this, we design two complementary comparisons that progressively isolate the effect of reasoning.

We first compare reasoning models against their chat counterparts within the same model family. As shown in~\autoref{performance2_all}, Qwen3-30B-A3B-Thinking (40.57\%) outperforms Qwen3-30B-A3B-Instruct (36.72\%) by 4 percentage points, and Qwen3-4B-Thinking (33.45\%) exceeds Qwen3-4B-Instruct-2507 (28.60\%) by approximately 5 percentage points. While this comparison suggests reasoning benefits, it conflates reasoning capability with potential differences in training data or optimization objectives between model variants.

To isolate the pure effect of reasoning, we leverage hybrid models that support both thinking and non-thinking modes within identical architectures and parameters. As shown in~\autoref{fig:bar_plot} (right), thinking mode consistently outperforms non-thinking mode across all scales from 0.6B to 32B. Crucially, the reasoning advantage scales with model capacity: the gap expands from 0.52 percentage points at 0.6B to 9.47 percentage points at 32B. This scaling pattern reveals an important insight—larger models possess greater reasoning potential, but this potential remains latent unless explicitly elicited through chain-of-thought processes. The multi-statement nature of \benchmark amplifies this effect: as models must integrate judgments across diverse knowledge points, the cumulative benefit of step-by-step reasoning compounds, making explicit reasoning increasingly valuable at larger scales.

The 14B model presents an instructive anomaly: under non-thinking mode, it achieves lower accuracy (17.98\%) than both 8B (19.61\%) and 4B (18.18\%) models. Response analysis reveals that the 14B model frequently skips analytical steps, outputting only final answers with an average length of merely 40.73 tokens (see~\Cref{appendix:response_length_analysis}). This "lazy answering" pattern demonstrates that \benchmark inherently penalizes shortcuts—models attempting to bypass careful statement-by-statement analysis suffer significant performance degradation, confirming that our benchmark genuinely assesses comprehensive reasoning capabilities rather than rewarding superficial pattern matching.

\begin{table}[ht]
\centering
\caption{Overall performance of selected models on different aggregation levels.}
\label{tab:discipline_performance}
\small
\resizebox{\textwidth}{!}{%
\begin{tabular}{@{}lccccccc@{}}
\toprule
\textbf{Metric} & 
\thead{OpenAI\\GPT-5.1-high} & 
\thead{Gemini-3\\Pro-Preview} & 
\thead{Gemini-2.5\\Pro} & 
\thead{OpenAI\\o3-high} & 
\thead{Qwen3-235B\\-A22B-Instruct} & 
\thead{Kimi\\K2-Instruct} & 
\thead{DeepSeek-V3.1\\Term.-non-thinking} \\
\midrule
Sample-wise (5038) & 62.07 & 61.75 & 58.93 & 58.36 & 50.40 & 44.66 & 42.70 \\
Subfield-wise (62) & 63.92 & 61.78 & 59.61 & 59.75 & 51.19 & 44.24 & 42.70 \\
Field-wise (44) & 63.71 & 62.40 & 59.03 & 59.56 & 49.99 & 43.20 & 42.58 \\
Discipline-wise (11) & 64.97 & 63.41 & 61.30 & 59.72 & 51.02 & 44.23 & 42.59 \\
\bottomrule
\end{tabular}%
}
\end{table}

\textbf{Impact of Aggregation Granularity.} \autoref{tab:discipline_performance} presents the performance of selected models across four aggregation levels: sample-wise (5,038 questions), subfield-wise (62 subfields), field-wise (44 fields), and discipline-wise (11 disciplines). Most top-tier models exhibit higher scores when evaluated at coarser aggregation levels—OpenAI-GPT-5.1-high increases from 62.07\% (sample-wise) to 64.97\% (discipline-wise), and Gemini-2.5-Pro rises from 58.93\% to 61.30\%—suggesting that these models perform relatively better in subfields with fewer samples. In contrast, Kimi-K2-Instruct and DeepSeek-V3.1-Terminus-non-thinking display slight opposite trends, with Kimi-K2-Instruct declining marginally from 44.66\% to 44.23\%, which indicates potential gaps in their coverage of certain low-sample domains. Meanwhile, some models such as Qwen3-235B-A22B-Instruct (50.40\%–51.19\%) and DeepSeek-V3.1-Terminus-non-thinking (42.58\%–42.70\%) demonstrate remarkable stability across aggregation levels, reflecting more uniform capability distribution across knowledge domains. This variance across aggregation granularities underscores the importance of hierarchical structure of \benchmark, as sample-wise evaluation may inadvertently favor models excelling in data-rich domains, while discipline-wise evaluation provides a more balanced assessment by treating each knowledge area equally.

\subsection{Further Analysis}

\begin{figure*}[htbp]
\begin{center}
\includegraphics[width=0.47\linewidth]{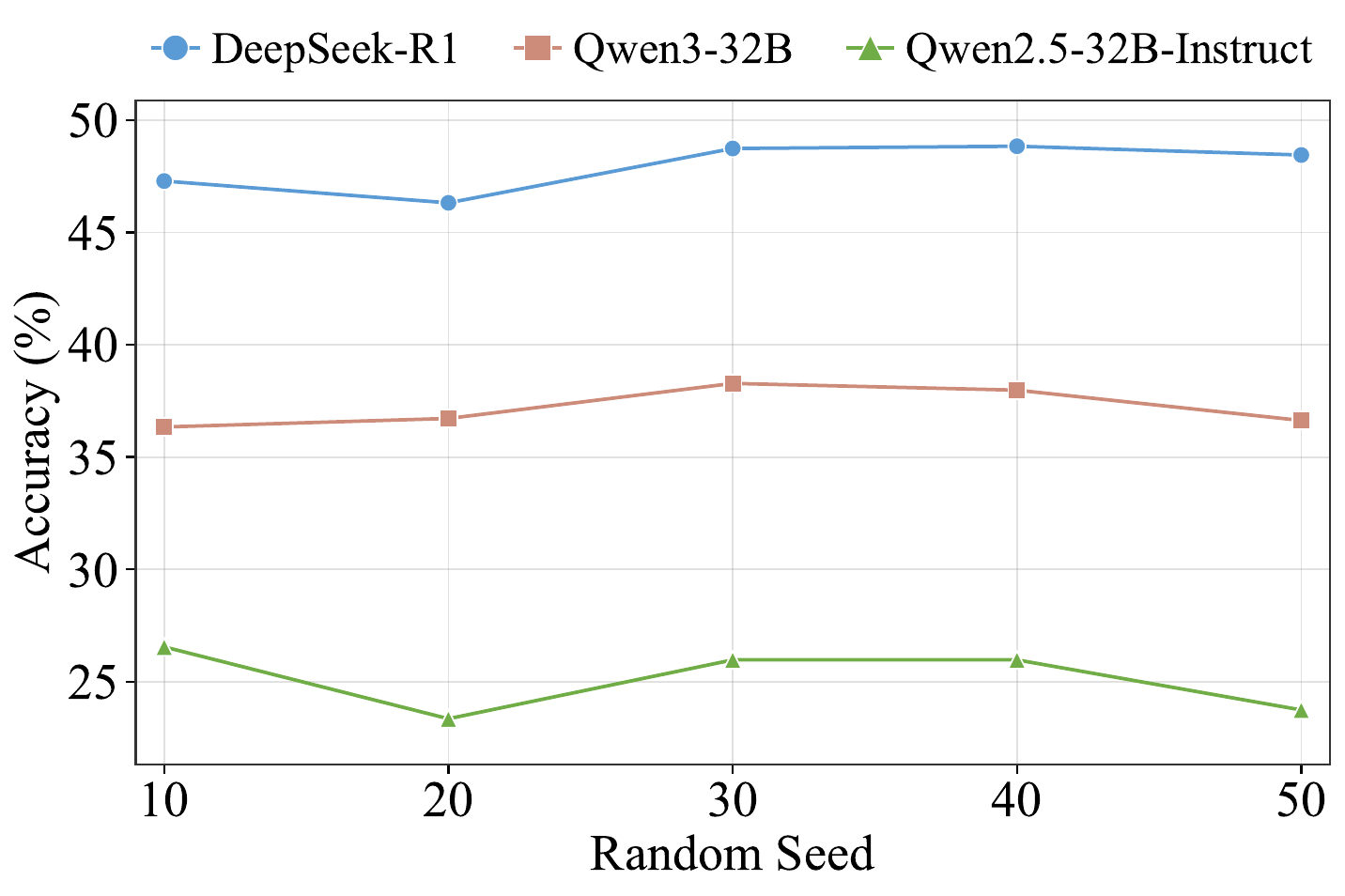}
\hspace{1em}
\includegraphics[width=0.47\linewidth]{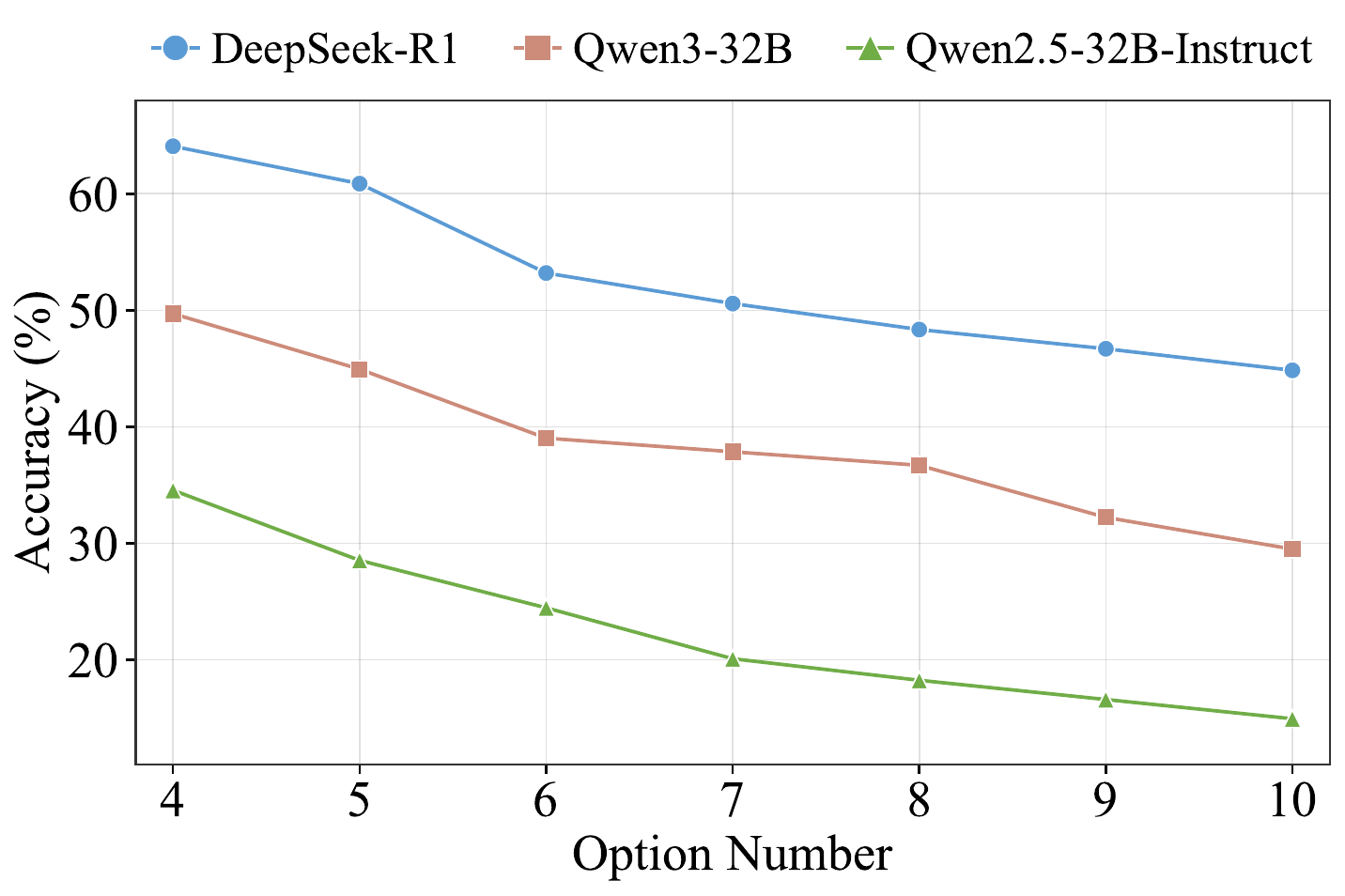}
\end{center}
\caption{Left: Model accuracy across five dynamically generated question sets with different random seeds.  Right: Impact of option count (4--10) on model accuracy.}
\label{fig:dynamic_analysis}
\end{figure*}

\textbf{Dynamic Evaluation.}
A key advantage of \benchmark is its ability to dynamically generate diverse question sets by varying random seeds that control statement selection and combination. To validate this property, we generate five distinct question sets and evaluate model performance across them. As shown in~\autoref{fig:dynamic_analysis} (left), two critical observations emerge. First, model rankings remain highly consistent across all seeds: DeepSeek-R1 consistently outperforms Qwen3-32B, which in turn surpasses Qwen2.5-32B-Instruct, with no rank reversals observed. Second, absolute performance fluctuations are minimal—DeepSeek-R1 varies within 46.32\%--48.84\%, Qwen3-32B within 36.34\%--38.28\%, and Qwen2.5-32B-Instruct within 23.35\%--26.55\%. \textbf{This stability, despite entirely different question compositions, confirms that our combinatorial design creates a vast question space resistant to memorization-based shortcuts.} Consequently, \benchmark enables periodic dataset refresh—releasing new question sets over time to prevent overfitting to any one fixed evaluation instance while maintaining reliable and comparable model rankings.

\textbf{Impact of Option Count.}
We further investigate how the number of candidate options affects model performance by varying the option count from 4 to 10. As illustrated in~\autoref{fig:dynamic_analysis} (right), increasing the number of options consistently degrades model accuracy across all evaluated models. DeepSeek-R1 demonstrates the strongest robustness, with accuracy declining from 64.08\% at 4 options to 44.85\% at 10 options. Qwen3-32B exhibits a similar trend, dropping from 49.71\% to 29.51\%. Qwen2.5-32B-Instruct shows the steepest decline, falling from 34.56\% at 4 options to merely 14.95\% at 10 options. This negative correlation between option count and accuracy is expected, as more options increase the cognitive load and reduce random guessing success probability. Notably, the performance gap between models widens as option count increases, suggesting that stronger models possess more robust knowledge reasoning capabilities that are less affected by increased task difficulty. These findings validate the flexibility of \benchmark in adjusting evaluation difficulty and provide insights into models' true capabilities beyond simple pattern matching.

\begin{table}[htbp]
\centering
\small
\caption{Accuracy (\%) across Option Positions A--H. Parentheses indicate total instances per position.}
\label{tab:position_bias}
\begin{tabular}{lcccccccc}
\toprule
Model & \thead{A\\(880)} & \thead{B\\(877)} & \thead{C\\(913)} & \thead{D\\(884)} & \thead{E\\(657)} & \thead{F\\(435)} & \thead{G\\(262)} & \thead{H\\(130)} \\
\midrule
Gemini-2.5-Pro & 55.11 & 59.98 & 62.54 & 62.56 & 55.86 & 56.09 & 53.05 & 64.62 \\
DeepSeek-R1    & 46.36 & 55.30 & 51.92 & 46.72 & 47.34 & 46.67 & 44.66 & 43.85 \\
Qwen3-32B      & 23.41 & 42.19 & 42.06 & 44.91 & 38.36 & 32.64 & 33.59  & 30.00 \\
\bottomrule
\end{tabular}
\end{table}

\textbf{Position Bias.}
A common concern in multiple-choice evaluation is whether models exhibit systematic biases toward specific option positions. As shown in~\autoref{tab:position_bias}, we compute accuracy distributions across positions A through H for three representative models. The results reveal no significant position bias: accuracy remains relatively stable across all positions for each model, with no consistent preference for earlier or later options. Gemini-2.5-Pro ranges from 53.05\% to 64.62\%, DeepSeek-R1 from 43.85\% to 55.30\%, and Qwen3-32B from 23.41\% to 44.91\%. The absence of position bias confirms that \benchmark provides fair evaluation, with model performance reflecting genuine reasoning rather than positional shortcuts.

\section{Limitations}
This work introduces a statement-based benchmark construction method featuring dynamic evaluation, multi-statement comprehensive understanding, and expert-free annotation, which together offer a cost-effective alternative to traditional approaches. While our current implementation demonstrates the feasibility and effectiveness of this method, several aspects remain to be explored in future work. Specifically, the current version of \benchmark covers only two languages (English and Chinese) and a constrained set of disciplines. Just as an encyclopedia is not built in a day, our low-cost, expert-free construction methodology is inherently designed for incremental expansion—new languages and disciplines can be seamlessly incorporated over time without requiring expensive re-annotation or domain expertise. We hope to build a multilingual, encyclopedic-scale evaluation framework that spans all world languages and knowledge domains, ultimately benefiting humanity as a whole and serving diverse industries and communities across the globe.

\clearpage

\bibliographystyle{plainnat}
\bibliography{main}

\newpage
\appendix

\crefalias{section}{appendix}
\crefalias{subsection}{appendix}
\section{Details of the Dataset Generation Pipeline}
\label{appendix:pipeline}

\newtcolorbox[auto counter, number within=section]{promptbox}[2][]{%
  colback=white, 
  colframe=purple!70!blue!80!black,  
  width=\textwidth,
  arc=2mm, 
  boxrule=0.5mm, 
  title={\normalsize\faCommentDots\hspace{0.5em}#2},
  breakable,
  fonttitle=\bfseries\Large, 
  fontupper=\small
  #1
}

\newtcolorbox[auto counter, number within=section]{correctexample}[2][]{%
  colback=white, 
  colframe=teal!80!green!80!black,  
  width=\textwidth,
  arc=2mm, 
  boxrule=0.5mm, 
  title={\normalsize\faCheckCircle\hspace{0.5em}#2}, 
  breakable,
  fonttitle=\bfseries\Large, 
  fontupper=\small, 
  #1
}

\newtcolorbox[auto counter, number within=section]{incorrectexample}[2][]{%
  colback=white, 
  colframe=teal!80!green!80!black,  
  width=\textwidth,
  arc=2mm, 
  boxrule=0.5mm, 
  title={\normalsize\faTimesCircle\hspace{0.5em}#2}, 
  breakable,
  fonttitle=\bfseries\Large, 
  fontupper=\small, 
  #1
}

\newtcolorbox[auto counter, number within=section]{questionexample}[2][]{%
  colback=white, 
  colframe=teal!80!green!80!black,  
  width=\textwidth,
  arc=2mm, 
  boxrule=0.5mm, 
  title={\normalsize\faGraduationCap\hspace{0.5em}#2}, 
  breakable,
  fonttitle=\bfseries\Large, 
  fontupper=\small, 
  #1
}

In this section, we provide a detailed breakdown of the prompts and templates utilized in our dataset construction pipeline. The process consists of three main stages: (1) extracting correct statements from images, (2) generating incorrect statements based on the correct ones, and (3) assembling the final question-answer pairs using a structured template. We present the specific prompts and corresponding examples for each stage below.

\subsection{Stage 1: Extracting Correct Statements}
\label{appendix:pipeline_stage1}
First, we employ an MLLM (doubao-1.5-vision-pro-32k) to analyze the input image and extract a set of grounded facts. These serve as the ground truth for the dataset.

\begin{promptbox}{Prompt: Correct Statement Extraction}
Image: \texttt{\{\{imageUrl\}\}}\\
Page Number: \texttt{\{\{Page Number\}\}}\\

1. Identify the knowledge point in the image. Note that the image contains only one complete knowledge point. Please provide the name and content of this knowledge point. Ensure the content is complete and retain as much information as possible. Keep formulas from the image unchanged and enclose them with \texttt{\$\$}. Place the extracted content in the designated location, specifically: \texttt{"content": "\{Knowledge Point Name\}: \{Knowledge Point Content\}"}.\\

Below is an example (including the original content and the concept to be extracted). Learn from this example and extract according to this standard.

\textbf{Original Content:}\\
ORDERED SETS:\\
1.5 Definition. Let $S$ be a set. An order on $S$ is a relation, denoted by $<$, with the following two properties:

(i) If $x\in S$ and $y\in S$, then one and only one of the statements $x < y$, $x = y$, $y < x$ is true.

(ii) If $x, y, z\in S$, if $x < y$ and $y < z$, then $x < z$.

The statement ``$x < y$'' may be read as ``$x$ is less than $y$'' or ``$x$ is smaller than $y$'' or ``$x$ precedes $y$''.

It is often convenient to write $y > x$ in place of $x < y$.

The notation $x\leq y$ indicates that $x < y$ or $x = y$, without specifying which of these two is to hold. In other words, $x\leq y$ is the negation of $x > y$.

1.6 Definition. An ordered set is a set $S$ in which an order is defined.

For example, $\mathbb{Q}$ is an ordered set if $r < s$ is defined to mean that $s - r$ is a positive rational number.

1.7 Definition. Suppose $S$ is an ordered set, and $E\subset S$. If there exists a $\beta\in S$ such that $x\leq\beta$ for every $x\in E$, we say that $E$ is bounded above, and call $\beta$ an upper bound of $E$.

Lower bounds are defined the same way (with $\geq$ in place of $\leq$).

1.8 Definition. Suppose $S$ is an ordered set, $E\subset S$, and $E$ is bounded above. Suppose there exists an $\alpha\in S$ with the following properties:

(i) $\alpha$ is an upper bound of $E$.

(ii) If $\gamma < \alpha$ then $\gamma$ is not an upper bound of $E$.

Then $\alpha$ is called the least upper bound of $E$ [that there is at most one such $\alpha$ is clear from (ii)] or the supremum of $E$, and we write $\alpha=\sup E$.

The greatest lower bound, or infimum, of a set $E$ which is bounded below is defined in the same manner: The statement $\alpha = \inf E$ means that $\alpha$ is a lower bound of $E$ and that no $\beta$ with $\beta > \alpha$ is a lower bound of $E$.

\textbf{Extracted Concept:}

Ordered Sets: An order on a set $S$ is a relation denoted by $<$, such that for any $x,y\in S$, one and only one of $x < y$, $x = y$, $y < x$ is true, and if $x,y,z\in S$, $x < y$ and $y < z$, then $x < z$. The notation $x\leq y$ means $x < y$ or $x = y$. For example, $\mathbb{Q}$ is an ordered set when $r < s$ is defined as $s - r$ being a positive rational number. Let $S$ be an ordered set and $E\subseteq S$. If there exists $\beta\in S$ such that $x\leq\beta$ for every $x\in E$, then $E$ is bounded above and $\beta$ is an upper bound of $E$; lower bounds are defined similarly (replacing $\leq$ with $\geq$). If $E$ is bounded above and there exists $\alpha\in S$ such that $\alpha$ is an upper bound of $E$ and if $\gamma < \alpha$, then $\gamma$ is not an upper bound of $E$, then $\alpha$ is called the supremum of $E$, denoted as $\alpha=\sup E$. The infimum of $E$, denoted as $\alpha = \inf E$, is defined similarly, meaning $\alpha$ is a lower bound of $E$ and no $\beta > \alpha$ is a lower bound of $E$. In the ordered set $\mathbb{Q}$, examples include the set $A$ (composed of all positive rational numbers $p$ such that $p^{2}<2$) which is bounded above but has no least upper bound, and the set $B$ (composed of all positive rational numbers $p$ such that $p^{2}>2$) which is bounded below but has no greatest lower bound. Also, for the set $E_1$ (composed of all $r\in\mathbb{Q}$ with $r < 0$) and $E_2$ (composed of all $r\in\mathbb{Q}$ with $r\leq0$), $\sup E_1=\sup E_2 = 0$, but $0\notin E_1$, $0\in E_2$; for the set $E$ (composed of $1/n$, $n = 1,2,3,\cdots$), $\sup E = 1$ ($1$ is in $E$) and $\inf E = 0$ ($0$ is not in $E$). An ordered set $S$ is said to have the least upper bound property if when $E\subseteq S$, $E$ is non-empty and bounded above, $\sup E$ exists in $S$. $\mathbb{Q}$ does not have this property, and an ordered set with the least upper bound property also has the greatest lower bound property. If $S$ is an ordered set with the least upper bound property, $B\subseteq S$, $B$ is non-empty and bounded below, and $L$ is the set of all lower bounds of $B$, then $\alpha=\sup L$ exists in $S$ and $\alpha=\inf B$.\\

2. The content in the column named \{Page Number\} represents the page location of this knowledge point. Replace the content in the \{page\_location\} column with the content from the \{Page Number\} column. For example: if the content in \{Page Number\} is 2, then the correct format should be \texttt{"page\_location":2}.\\

3. Language: Must be consistent with the image. If the image is in Chinese, the output should be in Chinese; if the image is in English, the output should be in English. Try not to alter the original expression; only extract.\\

Output Format: JSON format. Details as follows:

\begin{tcolorbox}[colback=gray!10, colframe=gray!20, boxrule=0.5pt, arc=0mm, top=2mm, bottom=2mm, left=2mm, right=2mm]
\ttfamily
\{ \\
\hspace*{1em} \{ \\
\hspace*{2em} "section\_order": "\{section\_order\}", \\
\hspace*{2em} "section\_title": "\{section\_title\}", \\
\hspace*{2em} "knowledges": [ \\
\hspace*{3em} \{ \\
\hspace*{4em} "page\_location": \{page\_location\}, \\
\hspace*{4em} "content": "\{content\}" \\
\hspace*{3em} \} \\
\hspace*{2em} ] \\
\hspace*{1em} \} \\
\}
\end{tcolorbox}

\end{promptbox}

\begin{correctexample}{Example Output: Correct Statements}
\texttt{"content":} "\textbf{Inverse Matrix:} Matrix inversion is a powerful tool in linear algebra for analytically solving equations. The inverse of matrix $A$ is denoted as $A^{-1}$, defined as the matrix such that $A^{-1}A = I_n$. An equation $Ax = b$ can be solved using $A^{-1}$ through steps: $A^{-1}Ax = A^{-1}b$, then $I_nx = A^{-1}b$, and finally $x = A^{-1}b$."
\end{correctexample}

\subsection{Stage 2: Generating Incorrect Statements}
\label{appendix:pipeline_stage2}
Based on the extracted correct statements, we prompt the model to generate plausible but false descriptions, referred to as incorrect statements. This step involves modifying key entities, attributes, or relationships to create challenging distractors.

\begin{promptbox}{Prompt: Incorrect Statement Generation}
You are a senior professor lecturing on the textbook \texttt{\{book\_name\}}, specializing in designing high-difficulty test questions to deeply assess outstanding students' mastery of theoretical knowledge. You will receive a correct statement enclosed in triple quotes ('''), and you need to transform it into a highly deceptive incorrect version.

\vspace{0.5em}
\textbf{Execution Steps:}
\begin{enumerate}
    \setlength{\itemsep}{0pt} \setlength{\parskip}{0pt} 
    \item \textbf{Analyze First:} Identify the core concepts of the correct statement and the best entry points for introducing errors.
    \item \textbf{Formulate Error Strategy:} Based on the analysis, select the most suitable strategy. Strategies include but are not limited to concept substitution, causal inversion, detail alteration, chronological displacement, logical gaps, and scope expansion/contraction.
    \item \textbf{Generate Incorrect Statement:} Generate a deceptive incorrect statement (The statement must maintain the academic tone and structure of the original text, appearing completely plausible and professional. Only students with deep understanding should be able to spot the issue. It must contain enough correct information to increase deceptiveness).
\end{enumerate}

\vspace{0.5em}
\textbf{The generated incorrect statement must meet the following quality standards:}
\begin{itemize}
    \setlength{\itemsep}{0pt} \setlength{\parskip}{0pt}
    \item \textbf{Professionalism:} Use accurate academic terminology and expressions.
    \item \textbf{Logical Coherence:} The surface-level logic must be complete with no obvious contradictions.
    \item \textbf{Deceptiveness:} Even outstanding students should need to think carefully to discover the problem.
    \item \textbf{Educational Value:} The error should reveal specific weak points in student understanding.
    \item \textbf{Consistency:} The generated incorrect statement should match the style, language, terminology, and expression of the correct statement.
\end{itemize}

\vspace{0.5em}
\textbf{Output the generated incorrect statement and reason strictly in the following JSON format:}

\begin{tcolorbox}[colback=gray!10, colframe=gray!20, boxrule=0.5pt, arc=0mm, top=2mm, bottom=2mm, left=2mm, right=2mm]
\ttfamily 
\{ \\
\hspace*{1em} "transformed\_content": "[Generated deceptive incorrect statement]", \\
\hspace*{1em} "incorrect\_reason": "[Explain the error type and the knowledge point tested]" \\
\}
\end{tcolorbox}

\vspace{0.5em}
\textbf{Input Correct Statement:} \\
\texttt{'''} \texttt{\{input\_statement\}} \texttt{'''}

\end{promptbox}

\begin{incorrectexample}{Example Output: Incorrect Statements}
\texttt{"transformed\_content":} "\textbf{Inverse Matrix:} Matrix inversion is a powerful tool in linear algebra for analytically solving equations. The inverse of matrix $A$ is denoted as $A^{-1}$, defined as the matrix such that $A^{-1}A = I_n$. An equation $Ax = b$ can be solved using $A^{-1}$ through steps: \textcolor{red}{$A^{-1}Ax = b$, then $I_nx = b$, and finally $x = b$.}" \\

\texttt{"incorrect\_reason"}: "Error Type: Detail Alteration. Key Concept: When solving a linear system using matrix inversion, the inverse matrix must be applied to both sides of the equation, i.e., $A^{-1}(Ax) = A^{-1}b$. This example incorrectly assumes that applying the inverse to the left side leaves the right side $b$ unchanged, ignoring the necessary transformation to $A^{-1}b$, which leads to the erroneous conclusion $x = b$."

\end{incorrectexample}

\subsection{Stage 3: Question Assembly}
\label{appendix:pipeline_stage3}

Finally, we assemble the correct and incorrect statements into a structured question format using a predefined template. In the template below, text highlighted in \textcolor{red}{\texttt{\{red braces\}}} denotes placeholders (e.g., \textcolor{red}{\texttt{\{option letter list\}}}) that are dynamically populated by program. 

\begin{promptbox}{Template: Question Assembly}
Answer the following multiple choice question. Please carefully analyze each statement and option, and end your response with 'Answer: \texttt{\$LETTER}' (no quotes), where \texttt{LETTER} is \textcolor{red}{\texttt{\{option letter list\}}}.

\vspace{1em}
\textcolor{red}{\texttt{\{statement list\}}}

\vspace{1em}
\textbf{Options:}\\
\textcolor{red}{\texttt{\{option list\}}}

\vspace{1em}
\hrule
\vspace{1em}

\textbf{predefined question template list:}

\vspace{0.5em}
\texttt{QUESTION\_TEMPLATES = [}
\begin{itemize}
    \setlength{\itemsep}{0pt} \setlength{\parskip}{0pt} \ttfamily \small
    \item[] "Given the following statements, which ones are \{accuracy\}? \{knowledge\_list\}",
    \item[] "Which of the following statements are \{accuracy\}? \{knowledge\_list\}",
    \item[] "From the statements below, identify the ones that are \{accuracy\}: \{knowledge\_list\}",
    \item[] "Please select all statements that are \{accuracy\}: \{knowledge\_list\}",
    \item[] "Among these statements, which ones are \{accuracy\}? \{knowledge\_list\}",
    \item[] "Review the following statements and identify those that are \{accuracy\}: \{knowledge\_list\}",
    \item[] "Choose all statements that are \{accuracy\} from the following: \{knowledge\_list\}",
    \item[] "Examine these statements and select the \{accuracy\} ones: \{knowledge\_list\}",
    \item[] "Which statements in the following list are \{accuracy\}? \{knowledge\_list\}",
    \item[] "Identify all \{accuracy\} statements from the following: \{knowledge\_list\}",
    \item[] "From the options below, select all \{accuracy\} statements: \{knowledge\_list\}",
    \item[] "Looking at these statements, which can be considered \{accuracy\}? \{knowledge\_list\}",
    \item[] "Read the following statements and determine which are \{accuracy\}: \{knowledge\_list\}",
\end{itemize}
\texttt{]}

\vspace{1em}

\texttt{ACCURACY\_TEMPLATES = \{}
\begin{itemize}
    \setlength{\itemsep}{0pt} \setlength{\parskip}{0pt} \ttfamily \small
    \item[] "correct": [
        "correct", "accurate", "true", "valid", "right", "factual", "precise", "exact", "proper", "sound"
    ],
    \item[] "incorrect": [
        "incorrect", "false", "invalid", "wrong", "inaccurate", "erroneous", "mistaken", "flawed", "faulty", "imprecise"
    ],
\end{itemize}
\texttt{\}}

\end{promptbox}

\begin{questionexample}{Example: Constructed Question}
Answer the following multiple choice question. Please carefully analyze each statement and option, and end your response with 'Answer: \texttt{\$LETTER}' (no quotes), where \texttt{LETTER} is A, B, C, D, E, F, G.

\vspace{0.5em}
From the statements below, identify the ones that are accurate:
\begin{enumerate}[label=\Roman*.]
    \setlength{\itemsep}{0pt} \setlength{\parskip}{0pt}
    \item Inverse Matrix: ...
    \item Eigenvalues and Eigenvector: ...
    \item Linear Dependence: ...
    \item Gibbs Sampling: ...
    \item The Trace Operator: ...
    \item Chain Rule of Probability: ...
    \item Diagonal Matrices: ...
    \item Bayes' Rule: ...
\end{enumerate}

\vspace{0.5em}
\textbf{Options:}
\begin{itemize}
    \setlength{\itemsep}{0pt} \setlength{\parskip}{0pt}
    \item[A)] IV, V, VIII
    \item[B)] I, II, III
    \item[C)] I, II, VIII
    \item[D)] I, III, VII
    \item[E)] I, III, VI
    \item[F)] I, II, V, VI
    \item[G)] I, IV
\end{itemize}
\end{questionexample}

\newpage
\section{Question distribution}
\label{appendix:question_distribution}
\begin{figure*}[ht]
    \centering  
    \includegraphics[width=\linewidth]{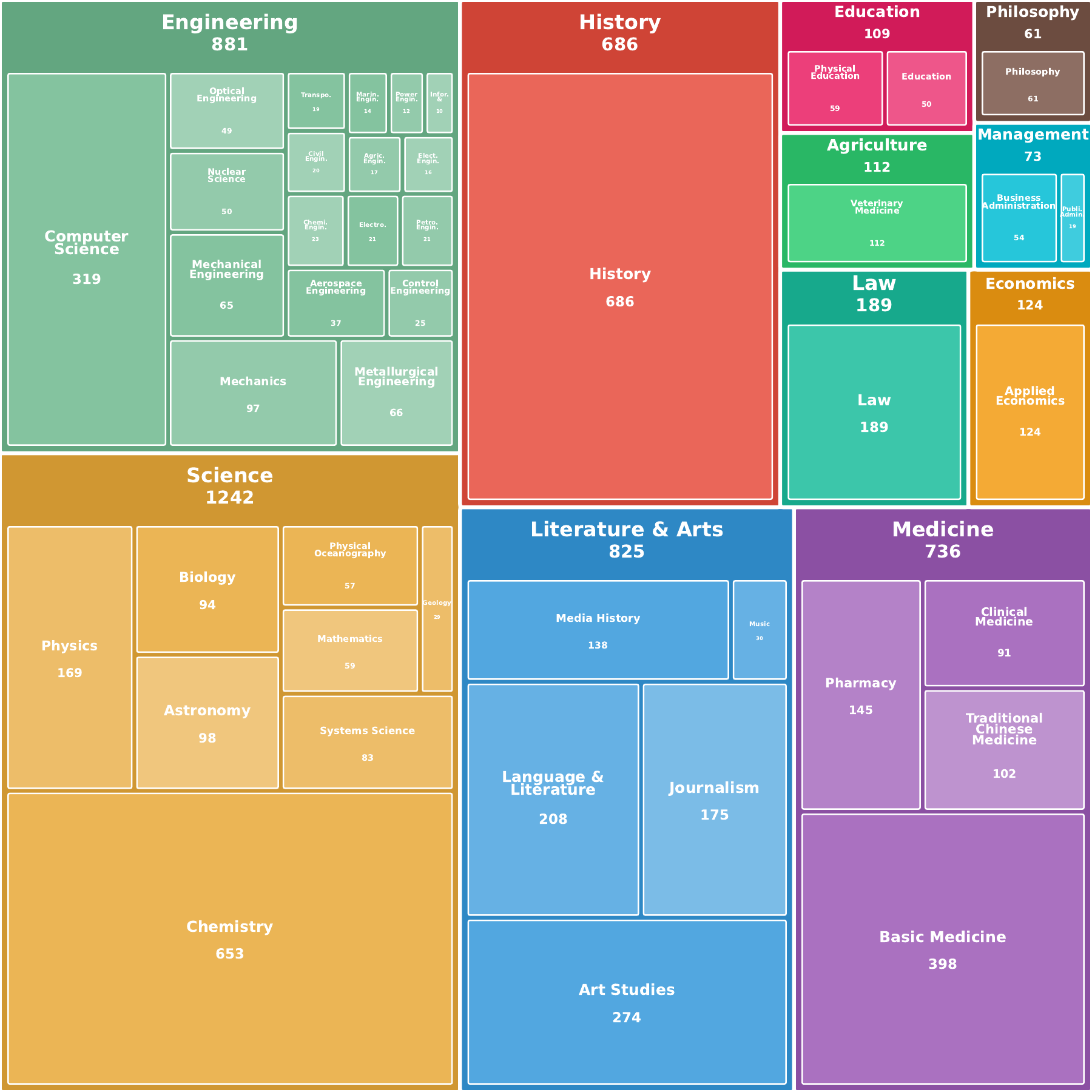}
    \caption{\textbf{\benchmark Question distribution.} Distribution of the Encyclo-K dataset, comprising 5,038 questions across 11 disciplines and 44 fields.}
    \label{fig:question_distribution}
\end{figure*}
\autoref{fig:question_distribution} shows the distribution of 5,038 questions across 11 disciplines and 44 fields. Engineering contains the most fields (18, e.g., Computer Science, Mechanics, Metallurgical Engineering), followed by Science (8), Literature and Arts (5), and Medicine (4). The remaining disciplines, such as History and Law, typically contain only one or two fields. This figure displays only discipline and field levels; subfields are omitted due to layout constraints. For example, the history discipline has one field (History), which encompasses two subfields: World History and Chinese History.

\section{Quality Control Details}
\label{appendix:quality_control}

\textbf{Correct Statement Construction.} We sequentially filter out statements containing images (3,269 instances), statements with chapter references, formula numbers, or chemical equations (642 instances), and statements with ambiguous semantics (127 instances), resulting in 21,525 correct statements. Three annotators manually review all correct statements and identify no significant issues.

\textbf{Incorrect Statement Construction.} Based on the correct statements, we use DeepSeek-R1 to rewrite them into incorrect counterparts and filter out overly short samples, yielding 21,494 incorrect statements. We randomly sample 200 instances for manual review and find that 5 samples have error reasons inconsistent with the annotations; however, the statements themselves remain factually incorrect, which does not affect the intended use of the dataset. Therefore, no additional processing is performed.

\textbf{Recommendations for Data Collection.} If the aforementioned issues are addressed during the data collection phase (e.g., masking formula numbers and chapter references), post-processing steps and review workload can be significantly reduced. We provide detailed data collection specifications in the main text (\Cref{subsubsec:collection}) for reference in future research.

\section{Multi-Statement Comprehension Analysis}
\label{appendix:multi_statement_analysis}

\begin{figure}[htbp]
\begin{center}
\includegraphics[width=0.8\linewidth]{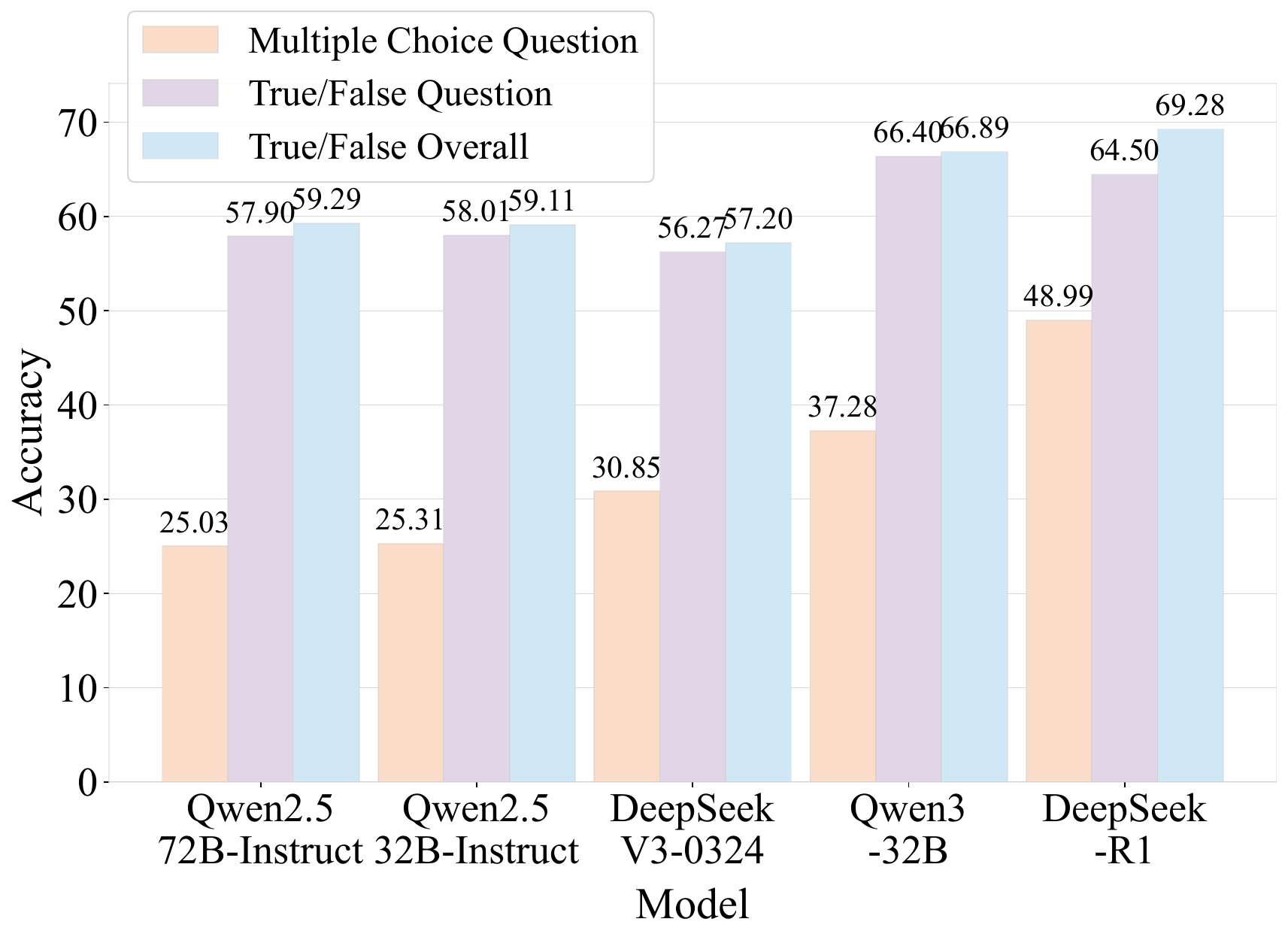}
\end{center}
\caption{Performance comparison across three evaluation settings: statement-level true/false judgment, question-level true/false judgment, and multi-statement comprehensive understanding.}
\label{fig:multi_statement_overall}
\end{figure}

As shown in~\autoref{fig:multi_statement_overall}, the main text compares single-statement judgment accuracy (averaged over the entire statement collection) with multi-statement comprehensive understanding accuracy to demonstrate the challenge introduced by \benchmark compared to previous benchmarks. We also report question-level true/false accuracy (purple), where each question's score is the average accuracy of judging its constituent statements. Even under this evaluation setting, model performance remains substantially lower than statement-level judgment, for example, DeepSeek-R1 drops from 69.28\% to 64.50\%, and further to 48.99\% on MCQ. This consistent performance degradation across evaluation settings confirms that \benchmark's multi-statement comprehensive understanding poses significant challenges beyond isolated statement verification.

\section{Response Length and Performance Analysis}
\label{appendix:response_length_analysis}

\begin{figure}[htbp]
\begin{center}
\includegraphics[width=0.7\linewidth]{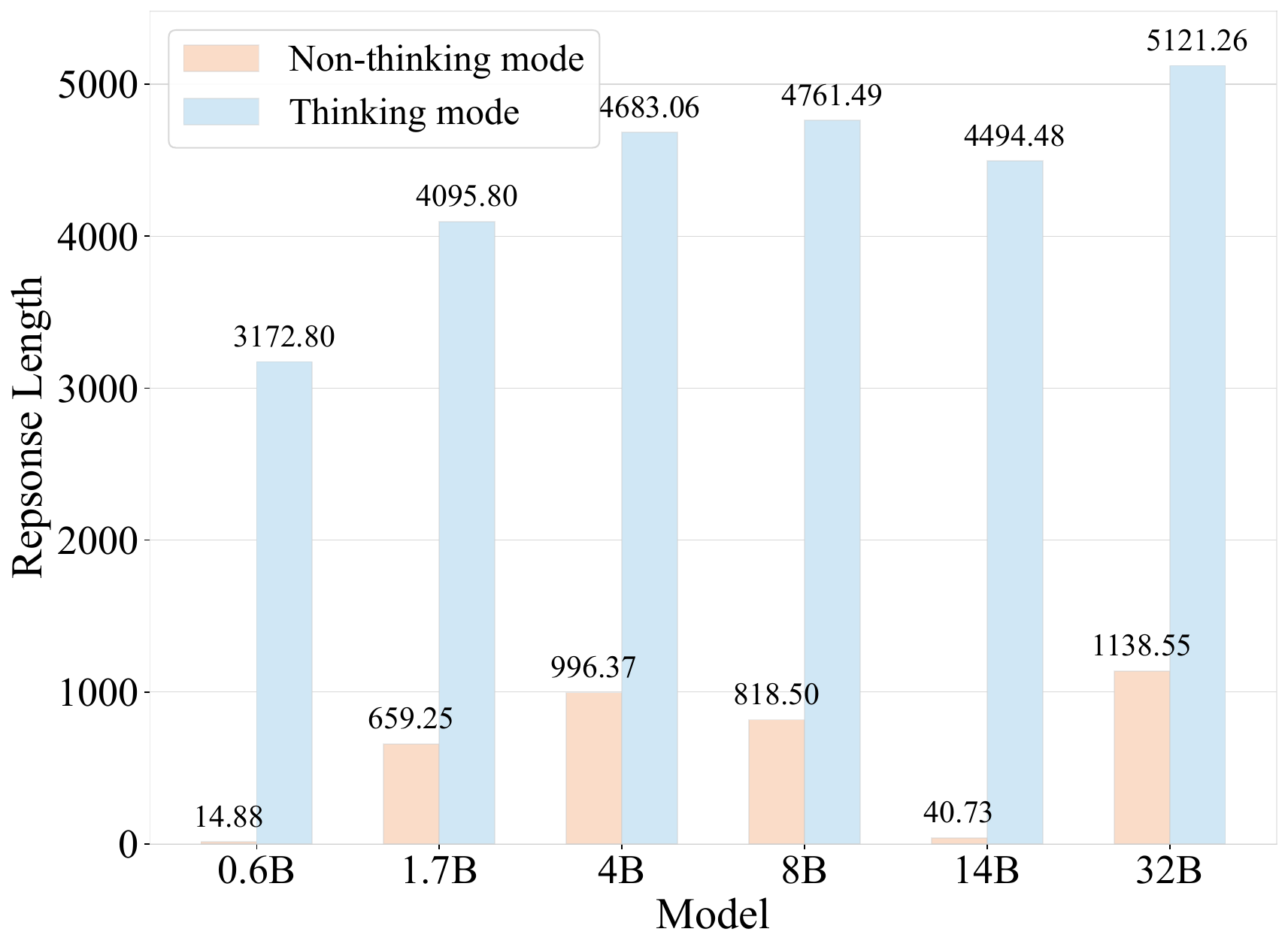}
\end{center}
\caption{Average response length (in tokens) of Qwen3 hybrid models under non-thinking and thinking modes. The 14B model exhibits anomalously short responses (40.73 tokens) under non-thinking mode.}
\label{fig:response_length_appendix}
\end{figure}

\autoref{fig:response_length_appendix} shows the average response length of Qwen3 hybrid models under thinking and non-thinking modes. Thinking mode produces substantially longer responses (3,172--5,121 tokens) compared to non-thinking mode (14--1138 tokens). Notably, the 14B model under non-thinking mode generates only 40.73 tokens on average—far shorter than the 8B (818 tokens) and 4B (996 tokens) models—which explains its anomalous performance degradation (17.98\% accuracy) discussed in the main text.

\end{document}